\documentclass{article} % For LaTeX2e

\usepackage{amsmath,amsfonts,bm}

\def\eqref#1{equation~\ref{#1}}
\def\1{\bm{1}}

\DeclareMathAlphabet{\mathsfit}{\encodingdefault}{\sfdefault}{m}{sl}
\SetMathAlphabet{\mathsfit}{bold}{\encodingdefault}{\sfdefault}{bx}{n}

\usepackage[hyphens]{url}
\usepackage{hyperref}
\usepackage[accepted]{icml2024}
\usepackage{times}

\hypersetup{colorlinks=true,breaklinks=true}
\usepackage{array} % Include the array package

\usepackage{graphicx}
\usepackage{framed}
\usepackage{subfigure}
\usepackage{amsmath}
\usepackage{comment}
\usepackage{hyperref}
\usepackage{array}
\usepackage{multirow}

\begin{document}

\twocolumn[
\icmltitle{Decoding In-Context Learning: Neuroscience-inspired Analysis of Representations in Large Language Models}

\icmlsetsymbol{equal}{*}

\begin{icmlauthorlist}
\icmlauthor{Safoora Yousefi}{Microsoft}
\icmlauthor{Leo Betthauser}{Microsoft}
\icmlauthor{Hosein Hasanbeig}{Microsoft Research}
\icmlauthor{Raphaël Millière}{Macquarie University}
\icmlauthor{Ida Momennejad}{Microsoft Research}
\end{icmlauthorlist}

\icmlaffiliation{Microsoft}{Microsoft, Redmond, WA}
\icmlaffiliation{Microsoft Research}{Microsoft Research, New York City, NY}
\icmlaffiliation{Macquarie University}{Macquarie University, Sydney, Australia}

\icmlcorrespondingauthor{Safoora Yousefi}{sayouse@microsoft.com}
\icmlcorrespondingauthor{Ida Momennejad}{idamo@microsoft.com}

\icmlkeywords{Large language models, in-context learning, representational similarity analysis, attention ratio analysis}

\vskip 0.3in]
\printAffiliationsAndNotice{}
\newcommand{\fix}{\marginpar{FIX}}
\newcommand{\new}{\marginpar{NEW}}

% use this macro to comment
\newcommand{\cmnt}[1]{\textcolor{blue}{[#1]}}
% use this macro for edits
\newcommand{\edit}[1]{\textcolor{red}{#1}}
% use this macro for todos
\newcommand{\todo}[1]{\textcolor{orange}{TODO: #1}}

\begin{abstract}
Large language models (LLMs) exhibit remarkable performance improvement through in-context learning (ICL) by leveraging task-specific examples in the input. However, the mechanisms behind this improvement remain elusive. In this work, we investigate how LLM embeddings and attention representations change following in-context-learning, and how these changes mediate improvement in behavior. We employ neuroscience-inspired techniques such as representational similarity analysis (RSA) and propose novel methods for parameterized probing and measuring ratio of attention to relevant vs. irrelevant information in Llama-2~70B and Vicuna~13B. We designed two tasks with a priori relationships among their conditions: linear regression and reading comprehension. We formed hypotheses about expected similarities in task representations and measured hypothesis alignment of LLM representations before and after ICL as well as changes in attention. Our analyses revealed a meaningful correlation between improvements in behavior after ICL and changes in both embeddings and attention weights across LLM layers. This empirical framework empowers a nuanced understanding of how latent representations shape LLM behavior, offering valuable tools and insights for future research and practical applications.
\end{abstract}

\section{Introduction}

Transformer-based large language models (LLMs) such as GPT-3 \citep{brown2020language} and Llama-2 \citep{touvronLlamaOpenFoundation2023} have achieved state-of-the-art performance on a wide range of tasks. One of the most intriguing aspects of modern Transformer-based models, especially LLMs, is their capacity for in-context learning (ICL) \citep{brown2020language}. ICL enables the model to improve its performance on new tasks from a few examples provided in the input context (or prompt), without any parameter updates. ICL enables LLMs to flexibly adapt their behavior to task-specific demands during inference without further training or fine-tuning. For instance, including examples of question-answer pairs in the prompt significantly improves performance on arithmetic, commonsense, and symbolic reasoning tasks \citep{weiChainofThoughtPromptingElicits2022,zhouTeachingAlgorithmicReasoning2022}. However, in spite of progress in this area, \textit{how} ICL improves behavior remains mostly unknown and an active area of research. %Here we propose a neuroscience-inspired approache to understanding ICL.

Some prior studies have framed ICL as implicit optimization, providing theoretical and empirical evidence that Transformer self-attention can implement algorithms similar to gradient descent  \citep{vonoswaldTransformersLearnIncontext2022,akyurekWhatLearningAlgorithm2022,ahnTransformersLearnImplement2023}. Other work has proposed a Bayesian perspective, suggesting pretraining learns a latent variable model that allows conditioning on in-context examples for downstream prediction \citep{xieExplanationIncontextLearning2022,wangLargeLanguageModels2023,ahujaInContextLearningBayesian2023}. While these formal investigations offer valuable insights, they generally focus on toy models trained on synthetic data that may fail to capture the full richness of ICL behavior exhibited by large models trained on internet-scale data. To advance a more complete understanding of ICL capabilities, new perspectives are needed to elucidate how ICL arises in LLMs trained on naturalistic corpora.

In this work, we introduce a neuroscience-inspired framework to empirically analyze ICL in two popular LLMs: Vicuna-1.3 13B \citep{vicuna2023}, and Llama-2 70B \citep{touvron2023llama}. We chose these models because they are open-source and provide access to embeddings and weights of all layers. We designed controlled experiments isolating diverse ICL facets including systematic generalization (e.g., in regression), and distraction resistance (e.g. in reading comprehension).

To interpret how the LLMs' latent representations support ICL, we focus on changes in the LLMs' embeddings and attention weights following ICL. We adopt methods from neuroscience, namely representational similarity analysis (RSA), to interpret how model representations in various hidden layers change as a result of ICL. We also propose Attention Ratio Analysis (ARA), a novel parameter-free method for computing the attention ratios between relevant and irrelevant information. Together, analyzing ICL-induced RSA and attention ratios enable us to systematically relate observed changes in latent representations and attention patterns to improvements in model behavior after ICL across three experiments. Our approach makes the following main contributions toward understanding the representational underpinnings of ICL.
\vspace{-2mm}
\begin{itemize}
\setlength\itemsep{1pt}
    \item We measured LLM embeddings for various prompts in the same task, and created hypothesis matrices about their expected similarity structure. We then measured the alignment of embedding similarities with the hypothesis using RSA. We found that ICL leads to changes in the prompt embedding similarity matrix to better encode task-critical information, which in turn improves behavior. We also decoded task-critical information prompt information from embeddings before and after ICL (using logistic regression classifiers) and showed that individual prompt embeddings also change after ICL to better encode this information. 
   
   \item We introduce attention ratio analysis (ARA) to measure changes in the attention weights after ICL. Specifically, we measure the ratio of attention between the models' response and the informative part of the prompt. We observed that increased allocation of attention to relevant content in the prompt correlates with behavioral improvements resulting from ICL.
   \item We performed RSA and ARA across different layers of various depths in Llama-2 and Vicuna and observed consistent patterns across layers.
\end{itemize}
\vspace{-2mm}

\emph{\textbf{A neuroscience-inspired approach.}} We use a technique known as representational similarity analysis (RSA), which is widely used across the neurosciences for comparison of neural representations and behavior. We believe RSA is suited to our approach for two reasons. First, parameter-free methods like RSA offer notable benefits over parameterized probing classifiers routinely used to decode internal model representations \citep{belinkovProbingClassifiersPromises2022}. This is because parameterized probes run a greater risk of fitting spurious correlations or memorizing properties of the probing dataset, rather than genuinely extracting information inherent in the representations \citep{belinkovProbingClassifiersPromises2022}. RSA avoids this risk, since it directly uses model activations. Second, unlike causal interventions used in mechanistic interpretability research~\citep{nanda2023progress, conmy2023towards}, RSA also scales efficiently to very large language models. Note that there is a potential trade-off between accuracy and complexity in probing methods, e.g., parameterized probes have the potential to achieve higher accuracy by fitting more complex patterns. In this work, we combine parameterized and parameter-free methods to leverage their respective strengths.

This work demonstrates the promise of neuroscience-inspired analysis in advancing interpretability and design of robust, capable LLMs.

\section{Methods}
\begin{figure*}[!t]
\centering
    \includegraphics[width=.8\textwidth]{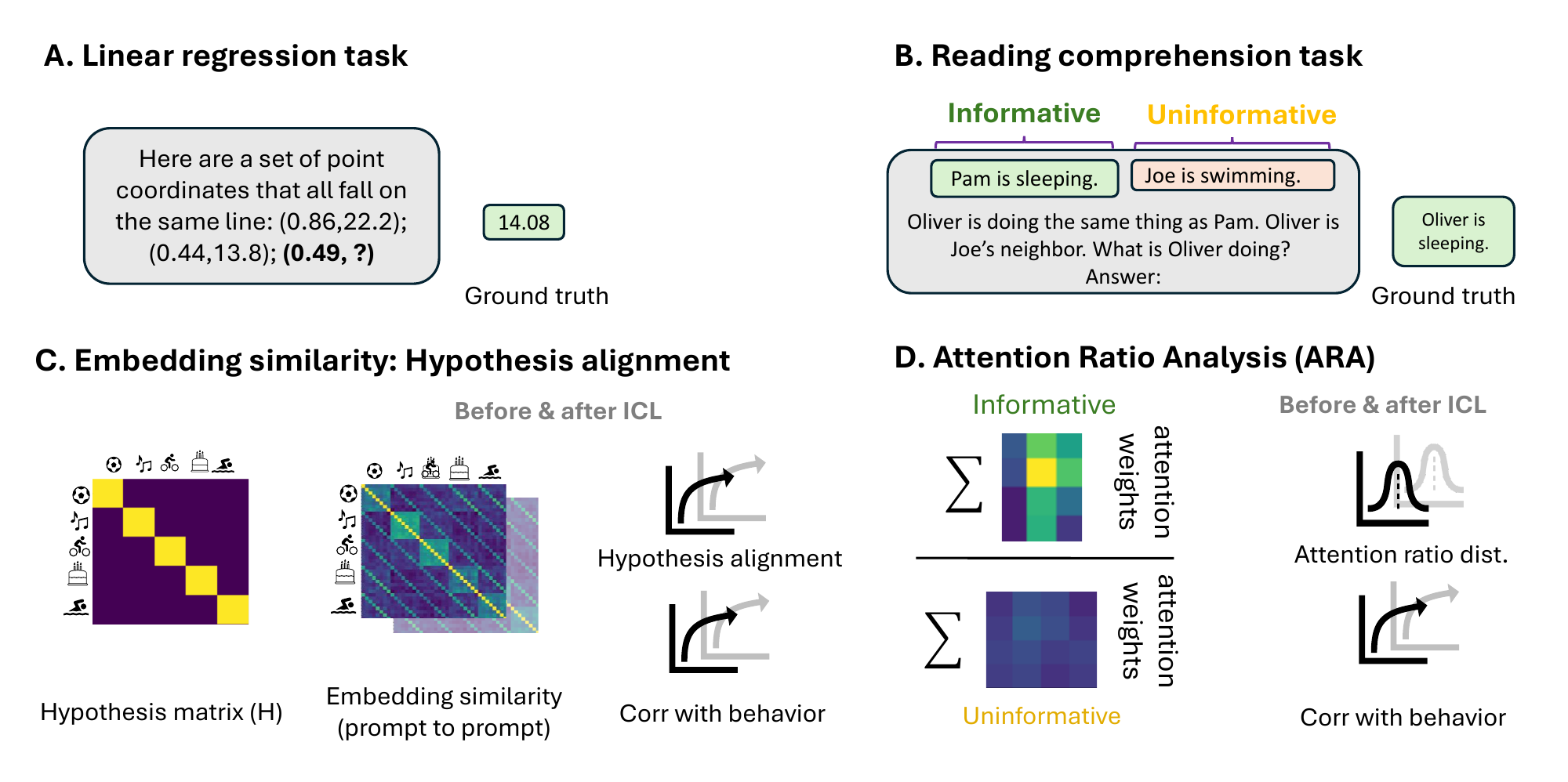}
\caption{\textbf{Experimental tasks and analyses.} (a) We designed a linear regression task, in which the LLM is provided with a set of $x$ and $y$ coordinates that fall on the same line, is given a final $x$ and is asked to provide a $y$ so all points fall on the same line. (b) We designed a reading comprehension task with prompts about individuals doing different activities and a question about the activity of one of the individuals. Crucially, the prompts included both informative and distracting subsequences. (c) We constructed a hypothesis matrix about the similarity of different prompts for a given task. Regression: hypothesis was based on the line's slope. Reading comprehension: three hypothesis matrices were constructed based on name, activity, and their combination corresponding to the correct response. We demonstrated the correlation of the alignment between the hypothesis and the embedding similarity matrices with LLM behavior. (d) We computed the ratio of attention to informative components of the prompt before and after ICL and its correlation with LLM behavior.}
\end{figure*}

In what follows we briefly describe the tasks, and the different steps of the latent representation analysis. This includes obtaining embedding vectors, performing representational similarity analysis (RSA) and attention ratio analysis (ARA), and training embedding classifiers.

\textbf{\emph{Tasks.}} We design two categories of experimental tasks throughout the paper. First, we demonstrate RSA as a tool to decode LLMs' hidden states in a linear regression (Section \ref{regression_exps}) task where a set of points from a line are given to the LLM and it is asked to generate another point on the same line. Then, we propose ARA as a tool to evaluate attention weights in a reading comprehension task (Section \ref{Exp1}), where the LLMs are given a short text including informative and distracting content, and are asked a question about the text.

\subsection{Obtaining Prompt Embeddings and Attention Weights}
\label{emb-proc}
The first step to perform RSA on LLM representations is to compute an embedding vector for each prompt. To this end, after feeding the prompt to the LLM, we first extract the $n$ $d$-dimensional embeddings for $n$ prompt tokens from the desired layer of the LLM. Then, we aggregate the token embeddings to reduce them into one $d$-dimensional vector per prompt. We experimented with both max-pooling and mean-pooling and verified that our findings are robust to the choice of embedding aggregation method. Following \citet{barknobite}, we then standardize the embeddings before computing pairwise cosine similarities to reduces the effect of rogue dimensions dominating the similarities. We use the resulting embeddings as described in Sections \ref{sec:meth_rsa} and \ref{subsection:emb_class}.

For attention ratio analysis, we require the attention matrix between the response and the prompt. To obtain this, we first feed the concatenation of prompt and model response back to the model, then extract attention matrices from all $h$ attention heads in each desired layer. We then aggregate the attention weights from the $h$ attention heads into a single matrix for each prompt-response pair. We tried both max-pooling and mean-pooling to verify that our findings are robust to the choice of aggregation method.

\subsection{Embedding Representational Similarity Analysis}
\label{sec:meth_rsa}
We designed a set of prompts with common components, such that representing those components is crucial in solving the prompt tasks. We hypothesize that in order to complete the tasks correctly, the LLMs must map prompts with similar components to similar representations. 

To verify this, we obtained embeddings for each prompt as described in Section \ref{emb-proc}, and calculated cosine similarities for all pairs of prompts ($M$). We compared the resulting embedding similarity matrix with an a priori hypothesis matrix ($H$) representing the common components among prompts, arriving at a metric that we call \emph{hypothesis alignment}. Subsequently, we measured whether and how these representations and their hypothesis alignment change after we introduce ICL examples to the same prompts.

% \subsubsection{Embedding Similarity Concordance Index}
% \textcolor{orange}{TBD we may not need this metric in methods}
\subsection{Embedding Probes}
\label{subsection:emb_class}
While RSA allows studying the similarity structure among prompt embeddings, probing allows more targeted investigation of whether task-relevant features are encoded in them. As the prompt embeddings capture the LLMs' latent representations of the tasks, it is possible to decode various components of the task from prompt embeddings (Section~\ref{emb-proc}). We use a probing classifier to study the effect of in-context-learning on the decodability of task components of interest from the embeddings. To achieve this, we divide the prompts into training and testing sets, train a logistic regression model to predict a task component in the prompt, and measure the decoding accuracy on the test set. For example, in Section \ref{regression_exps}, each prompt involves a line, and the slope of the line is decoded from the prompt embeddings.  We report classification accuracy with 10 repetitions of Monte Carlo cross validation.

\subsection{Attention Ratio Analysis (ARA)}
\label{subsection:attn}
We examine representational changes in attention weights in various layers of the LLMs due to ICL. Namely, we compute the ratio of the attention between the tokens associated with the response and the tokens associated with different parts of the prompts. Using ARA, we example whether and increased attention to informative parts of the prompt underlies ICL.  

Intuitively, ARA measures how much of the response's attention is focused on a substring of interest in the prompt. Let's take $a$, $s$, and $t$ to denote three subsets of substrings contained in $x$ which represents the prompt concatenated with the response. For example, $a$ can be the prompt, $s$ the response, and $t$ can be a part of the prompt which contains the necessary information to answer the question. To enable measuring attention ratios, we construct an input string $x = p \frown r$ by concatenating the prompt $p$ and the model response $r$, and obtain the attention weights resulting from this input from the last layer of the model. Let $A$ be the aggregated attention weights across all of the desired layer's attention heads corresponding to $x$. We define the attention ratio $A(a,s,t)_{x}:=\frac{1}{|t(x, s)|}\Sigma(A_{i, j}) / \frac{1}{|t(x, t)|}\Sigma(A_{i, k})$ where $i \in t(x, a)$, $j \in t(x, s)$, $k \in t(x, t)$, and $t(u, v)$ indicates the token indices corresponding to substring $v$ after tokenization of string $u$. 

The attention ratio measure can be used to compare the attention of the response to relevant vs. irrelevant information in the prompt. We compared the distribution of this ratio before and after ICL in sections \ref{Exp1} across various Layers of Llama-2 and Vicuna. Two-sample t-test was used to measure the statistical significance of the shift in attention ratios before and after ICL.

\section{Linear Regression}
\label{regression_exps}
We first apply RSA to LLMs in the task of linear regression where we perform an in-depth study of the effect of number of in-context examples on LLMs' representation of lines. We took inspiration from \cite{coda2023meta} in designing this task. We created 256 prompts, and 16 different lines. In each prompt, we provided two to eight $(x_i, y_i)$ points for in-context learning and a test $x_T$ coordinate in the prompt, asking the LLM to predict the associated $y_T$ coordinate on the same line. The minimum number of points to determine a line is two, meaning we provide 0 to 6 additional examples for ICL. A sample prompt for this task with 1 additional example point is shown below:

\begin{framed}
{\small Here are a set of point coordinates that all fall on the same line: (0.86,22.2); (0.44,13.8); (0.63,17.6); (0.49,}
\end{framed}

\begin{figure*}[!t]
\centering
    \subfigure[Behavioral Error]{
        \includegraphics[width=.315\textwidth]{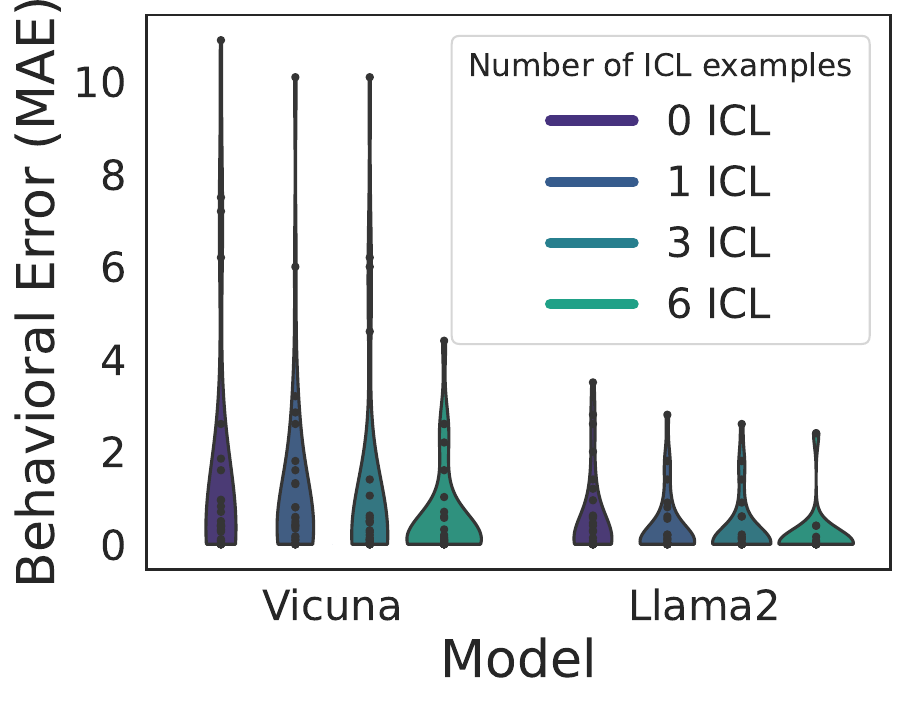}
        \label{sub_fig:reg_beh}} 
    \subfigure[Embedding Probe]{
        \includegraphics[width=.32\textwidth]{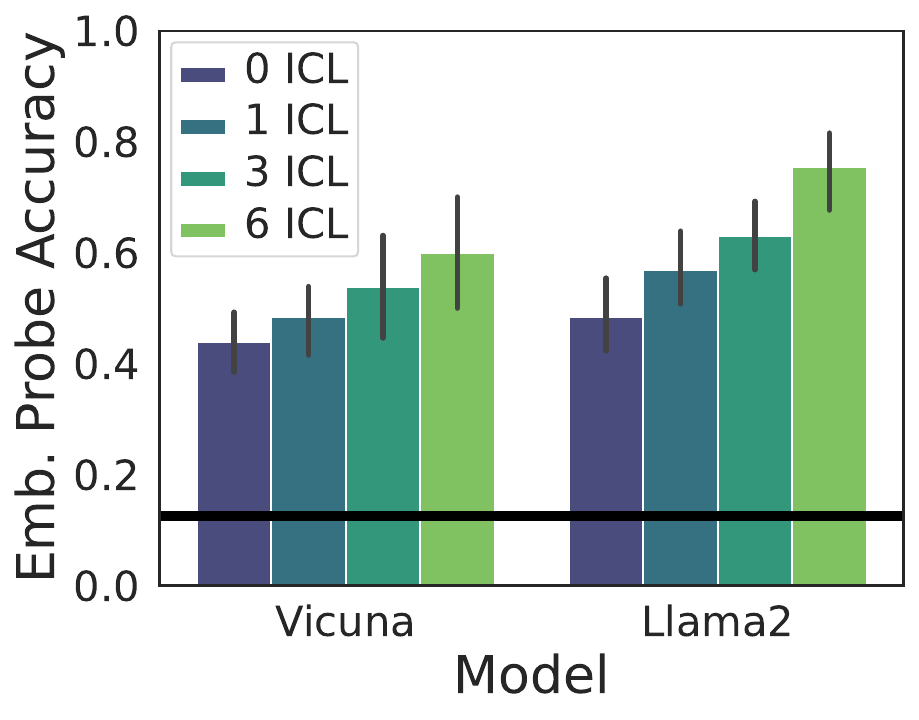}
        \label{sub_fig:reg_emb_probe}}
    \subfigure[Probe accuracy vs. behavior]{
        \includegraphics[width=.31\textwidth]{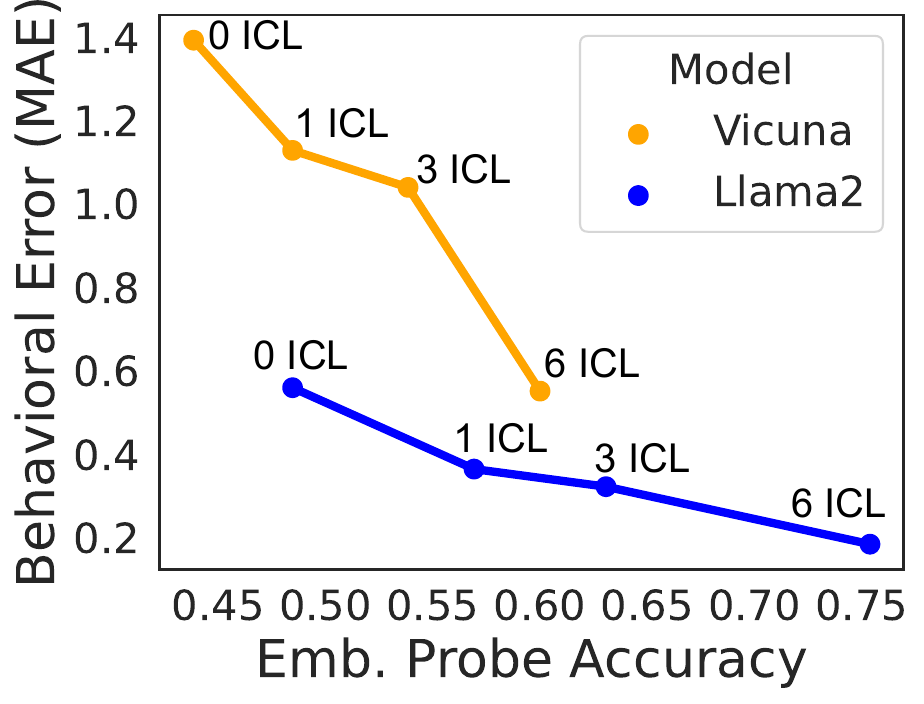}
        \label{fig:reg_emb_beh_corr}
        }
    \subfigure[Correlation(H, M)]{
    \label{fig:corr_reg}
    \includegraphics[width=.3\textwidth]{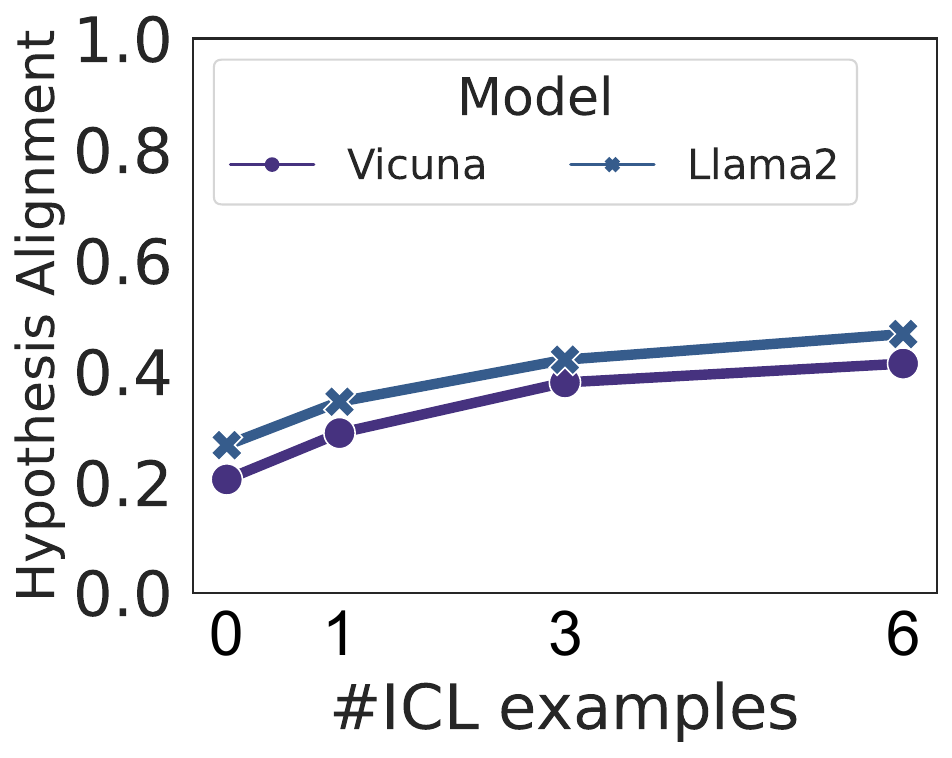}
    \label{fig:regression_rsa_beh}
    }
    \subfigure[Vicuna H-M alignment across layers]{
    \label{fig:corr_layers_vicuna}
    \includegraphics[width=.3\textwidth]{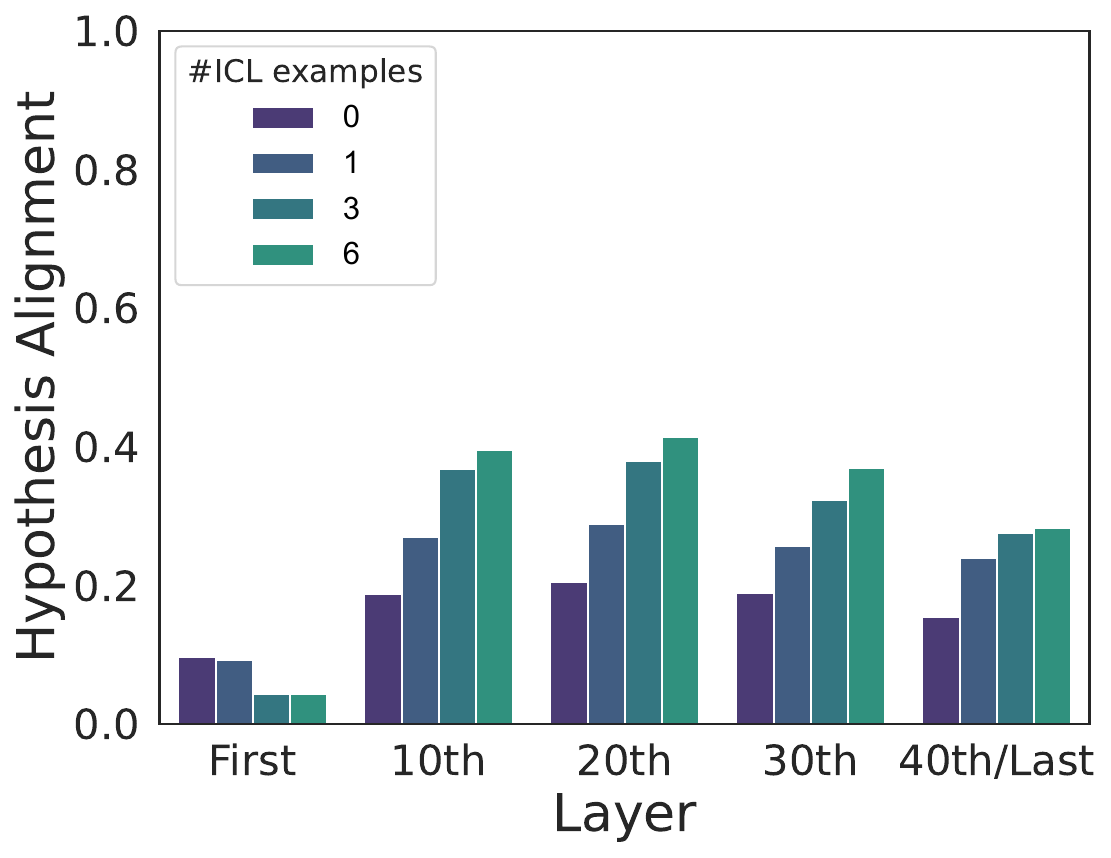}
    \label{fig:regression_rsa_layers_vic}
    }
    \subfigure[Llama2 H-M alignment across layers]{
    \label{fig:corr_layers_vicuna}
    \includegraphics[width=.3\textwidth]{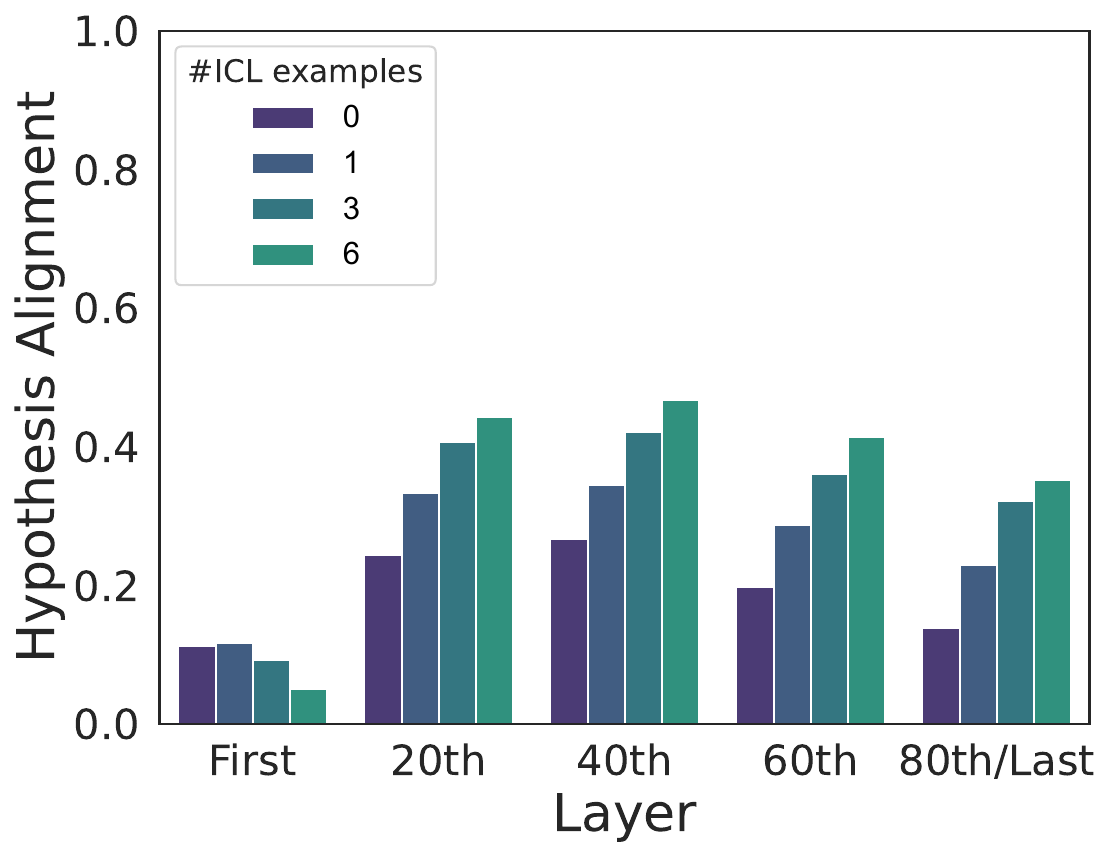}
    \label{fig:regression_rsa_layers_llama}
    }
    
    \caption{\textbf{Behavior, hypothesis alignment, and embedding probes for the regression task before and after ICL.} \textbf{(a)} Increasing the number of ICL examples decreased the absolute error between model's response $\hat{y_T}$ and the ground truth $y_T$ (see equation \ref{eq1}) for Llama-2 and Vicuna-1.3. \textbf{(b)} A logistic regression probing classifier was trained to predict the line slope of regression prompts from the last layer's embedding. Decoding accuracy increased with ICL in both Llama2 and Vicuna. \textbf{(c)} Behavior improvement in both models is correlated with the accuracy of the embedding classifier. The more slope information embedded in the model's representations, the smaller the model's mean absolute error in predicting $y_T$. \textbf{(d)} The correlation between our slope-based hypothesis matrix and the embeddings similarity matrix increases with more ICL examples for both models (visualised for middle layer). \textbf{(e) and (f)} Hypothesis alignment improved consistently with more ICL examples across LLM layers of varying depths with the exception of the first layer. }
    \label{fig:reg_rsa_results}
\end{figure*}

 We evaluated the models' behavioral error in calculating $y_T$ for $x_T$ as the absolute error between the response $\hat{y_T}$ and the ground truth:
\begin{equation}
    e_{reg}(y_T, \hat{y_T}) = | y_T - \hat{y_T}|
\label{eq1}
\end{equation}
Figure \ref{sub_fig:reg_beh} shows that the behavioral error of both models decreases as we increase the number of ICL examples. We studied representational changes in various layers of both LLMs that correlate with this behavioral change.
\begin{figure*}[!t]
\centering
    \subfigure[Slope-based Hypothesis H]{
    \label{fig:hyp_reg}
    \includegraphics[width=.25\textwidth]{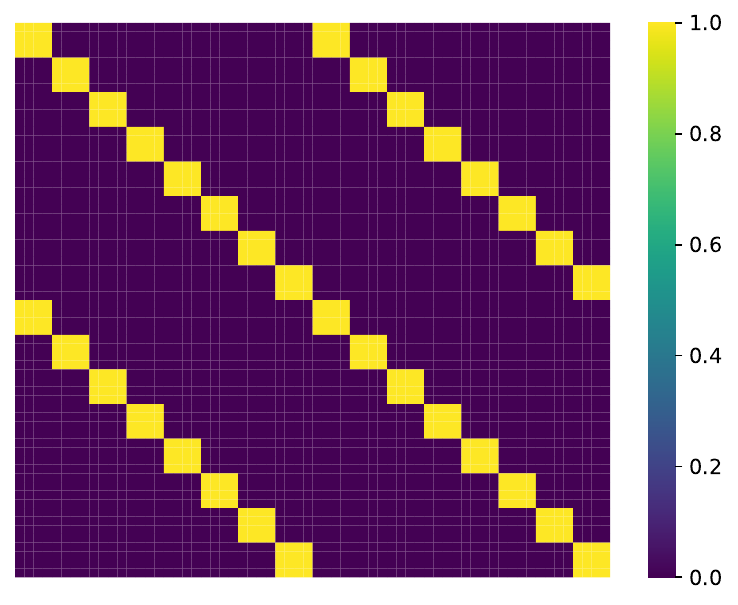}
    \label{fig:regression_rsa_a}
    }
    \subfigure[Embedding similarity $M_0$]{
    \label{fig:M_2_reg}
    \includegraphics[width=.25\textwidth]{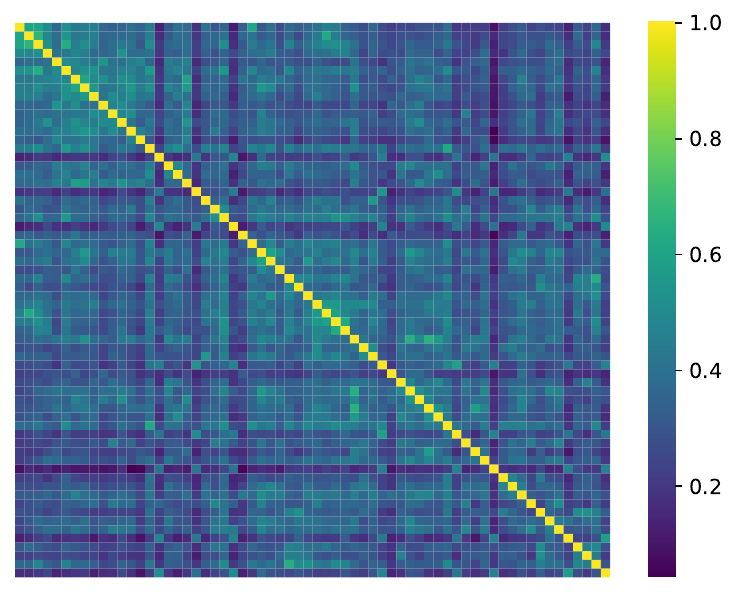}
    \label{fig:regression_rsa_b}
     }
    \subfigure[Embedding similarity $M_6$]{
    \label{fig:M_8_reg}
    \includegraphics[width=.25\textwidth]{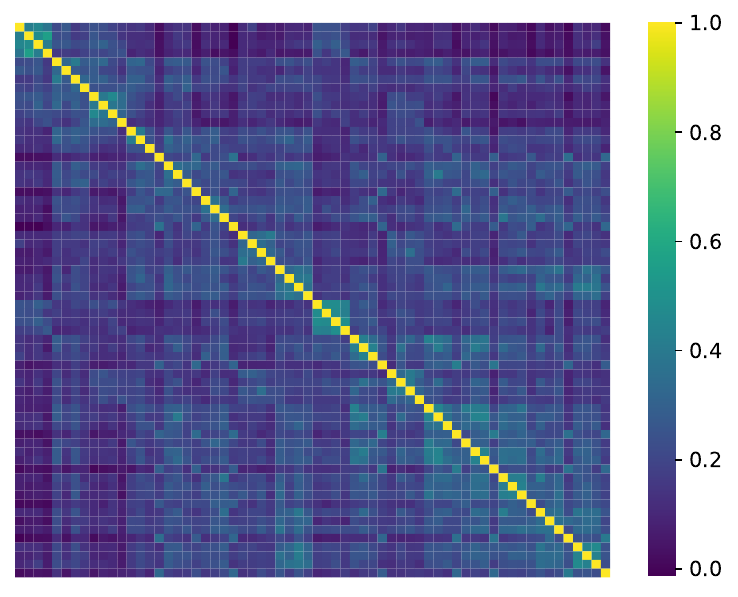}
    \label{fig:regression_rsa_c}
    }
    \caption{\textbf{Embedding similarity (M) and hypothesis (H) matrices for the regression task.} \textbf{(a)} We constructed a hypothesis similarity matrix assuming prompts about lines that have the same slope would have similar embeddings. \textbf{(b) and (c)} We computed the actual prompt-to-prompt embedding similarity matrix (for a given layer, e.g., the last layer of Llama-2 here) for prompts with no ICL ($M_0$), and after the addition of ICL examples ($M_k$). Each row and column represent the embedding of a regression task prompt. We then computed the alignment of the H and M similarity matrices before and after ICL for multiple layers of each model (Figure \ref{fig:reg_rsa_results}).}
    \label{fig:regression_rsa}
\end{figure*}

\subsection{Linear Regression RSA.} One underlying factor that could facilitate this behavioral improvement is better representation of the line's slope in the models' hidden states. Specifically, we hypothesize that prompts describing lines with the same slope should be encoded to similar embeddings in the models' representation space compared to those with different slopes. We built a similarity matrix $H$ to capture this hypothesis (Figure~\ref{fig:hyp_reg}), where the entry $H_{i,j}$ is 1 if the lines in the $i$th and $j$th prompt have the same slope and 0 otherwise. 

Next, we calculated the cosine similarity matrix among embeddings of prompts with $k$ ICL examples and we denoted it as $M_k$ (Figures~\ref{fig:M_2_reg} and \ref{fig:M_8_reg}). Finally, we computed \emph{hypothesis alignment} of the embeddings of prompts with varying numbers of ICL examples as the correlation between $H$ and each $M_k$. 

As shown in Figure~\ref{fig:corr_reg}, hypothesis alignment increased as we increased the number of ICL example points in the prompt for both models. Embeddings from the middle layer of both models were used in Figure~\ref{fig:corr_reg}, while we repeated this analysis using embeddings from five different layers of various depths and observed a similar increase in hypothesis alignment with all other layers except the first layer (Figures \ref{fig:regression_rsa_layers_vic} and \ref{fig:regression_rsa_layers_llama}).
We  repeated this experiment using mean-pooling over prompt token embeddings instead of max pooling, and although correlations were generally lower, we consistently observed improvement in hypothesis alignment with more ICL examples (Appendix Fig. \ref{fig:app_reg_rsa}).
% \todo{(Point to Appendix for mean aggregation)}
%\label{subsec:lin_reg_rsa}

\subsection{Classifying Line Information from Embeddings.} The RSA results suggested that increasing the number of ICL examples increases the amount of slope information in the embedding similarity matrix. We then used probing classifiers to investigate if the same information can be decoded from prompt embedding directly. We trained a logistic regression classifier to predict the slope of the line (8 total slopes) from the prompt embeddings of five different layers of Llama-2 and Vicuna-1.3. 

Increasing the number of in-context examples improved the classification accuracy of the slope classifiers in all layers except the first layer. (See Figure \ref{sub_fig:reg_emb_probe} for the last layer and Appendix Figure \ref{fig:app_reg_probe_slope_layers} for all layers.) This suggests that with more examples, LLMs were able to represent the line slopes more accurately. Importantly, behavioral improvement in both LLMs was correlated with the accuracy of this probing classifier. In other words, as the slope information embedded in the model’s latent representations increased, the models had smaller behavioral error in predicting $y_T$ for $x_T$. (Figure \ref{fig:reg_emb_beh_corr}). We observed similar results using mean-pooling over token embeddings instead of max-pooling (Appendix Figure \ref{subfig:reg_prob_c}).
\label{subsec:regression_classifier}

\section{Reading comprehension: Names and Activities}
\label{Exp1}
We designed a simple reading comprehension task that consists of clear and distinct components with a priori similarities, such that we can measure how each component is reflected in LLM embeddings and attention weights. 

\textbf{Simple prompts.} Specifically, we created 100 prompts of the form ``{name} + {activity}", using 10 distinct names and 10 distinct activities. We refer to these prompts as ``simple prompts" since we subsequently combine them to create more complex ``composite prompts". Here is an example of a simple prompt:
\begin{framed}
    {\small Patricia is reading a book.}
\end{framed}
% Figure \ref{fig:atomic-heatmap} displays the pairwise representational similarity matrix (Section \ref{sec:meth_rsa}) for the simple prompts in Llama-2's embedding space. The structure of this similarity matrix shows high similarity among prompts that involve the same name or the same activity. Visualization of Llama-2's embeddings after dimensionality reduction (tsne) further reveals clustering in the embedding space based on what activity is described in the prompt (Figure \ref{fig:atomic-tsne}).

\textbf{Composite prompts.} Next, we created composite prompts using two simple prompts. These composite prompts involve a simple reading comprehension task that requires the model to process information from the relevant simple prompt while ignoring the irrelevant one to come up with the correct answer.
Here is an example composite prompt created with two simple prompts:
\begin{framed}
{\small Question: Patricia is reading a book. Joseph is swimming. Oliver is doing the same thing as Patricia. Oliver is a neighbor of Joseph. What is Oliver doing? Answer:}
\end{framed}
\textbf{Behavioral change after ICL:} The correct response to the above prompt needs to include ``Oliver is reading a book." or ``Reading a book." Note that the prompt includes distracting statements that are not one of our simple prompts, e.g., ``Oliver is a neighbor of Joseph", to make the task more challenging. Our goal was to study how ICL can improve LLMs' performance despite of distractors. We found that different distractors pose a challenge to Llama-2 and Vicuna-1.3: We provide examples of prompts used for each model in the supplementary materials (Section \ref{sec:appendix_names_acts_sample_prompts}).

We created another 100 simple prompts with a different set of names and activities and made composite prompts to use as ICL examples. Llama-2 and Vicuna-1.3 performances on this task are reported before and after in-context examples were introduced (Figure \ref{sub_fig:nah-beh-both}). We observe that adding an ICL example significantly improves the performance of both models. In the following subsections, we use these composite prompts to analyze representational and attention changes underlying this behavior improvement.

\subsection{Embedding similarity and hypothesis alignment.} We constructed a prompt-to-prompt embedding similarity matrix using cosine similarity of prompt embeddings obtained from the LLMs. (See section \ref{emb-proc} for details).

\begin{figure*}[!t] 
    \centering
    \subfigure[Behavioral Accuracy]{
        \includegraphics[width=.27\textwidth]{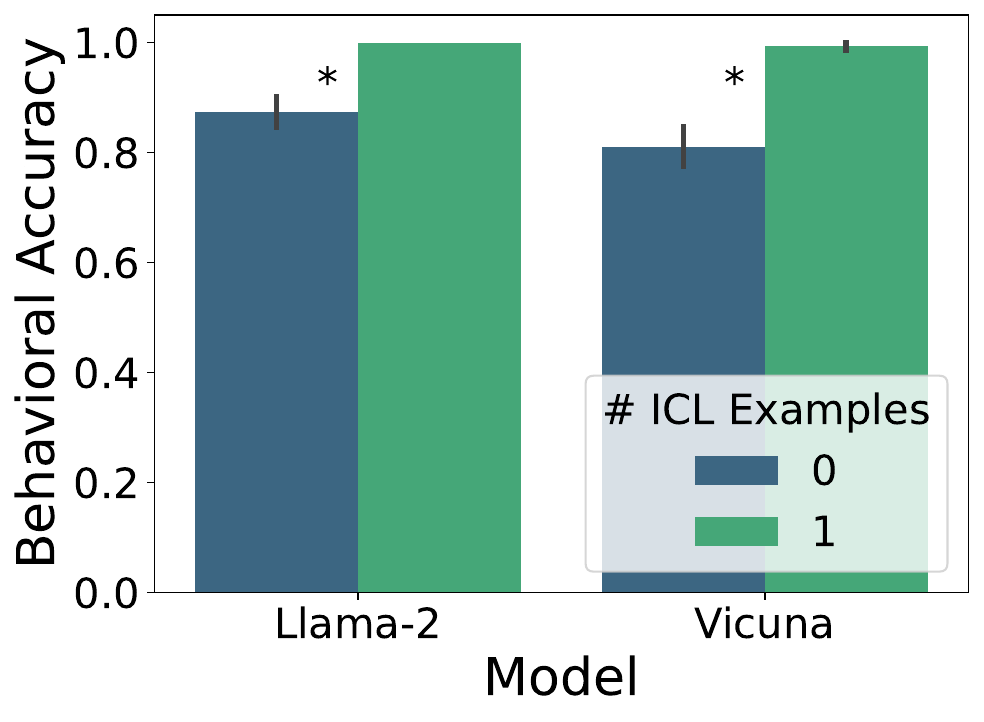}
        \label{sub_fig:nah-beh-both}}
    \hspace{6mm}
   \subfigure[Vicuna Hypothesis Alignment]{
        \includegraphics[width=.27\textwidth]{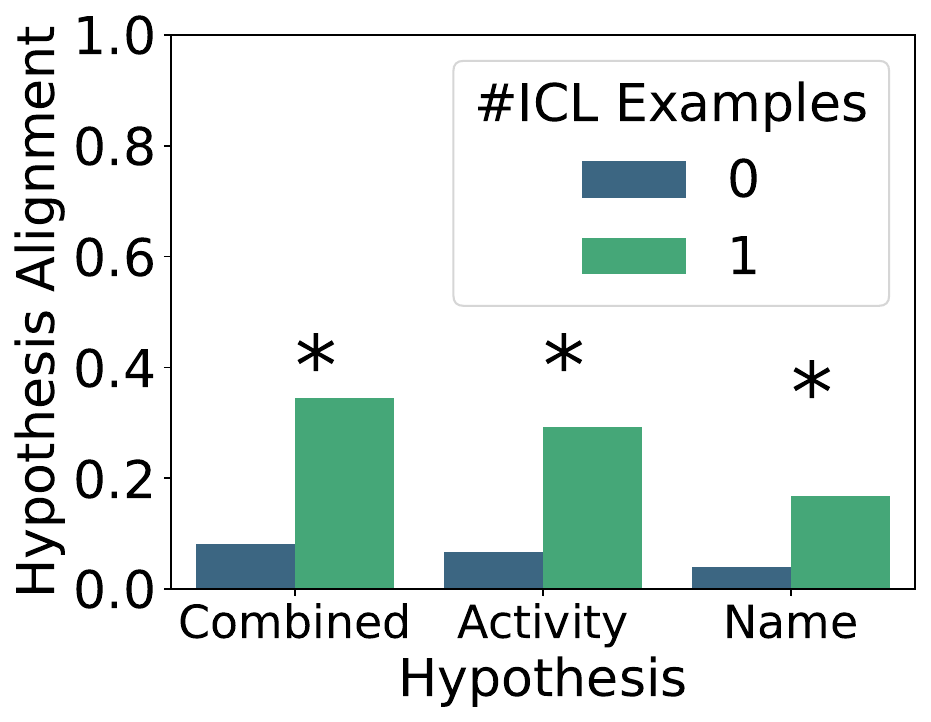}
        \label{sub_fig:nah-rsa-vicuna-all-hyp}}
    \hspace{6mm}
   \subfigure[Llama-2 Hypothesis Alignment]{
        \includegraphics[width=.27\textwidth]{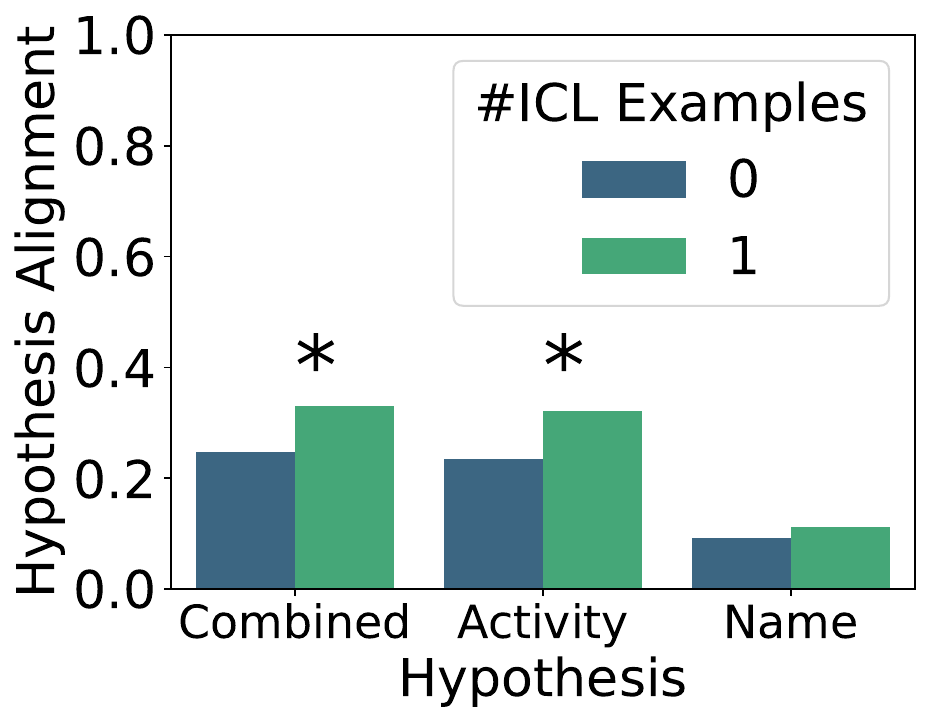}
        \label{sub_fig:nah-rsa-llama-all-hyp}}
    \caption{\textbf{Model behavior and hypothesis alignment before and after ICL for the reading comprehension task.} (a): The accuracy of Llama-2 and Vicuna-1.3 behavior in the reading comprehension task significantly benefits from ICL examples. (b) ICL significantly improves the alignment between Vicuna's embedding similarity matrix and three hypothesis matrices, $p< 0.05$: name-based similarity (prompts inquiring about the same individual are more similar), activity-based similarity (prompts whose correct answer includes the same activity are more similar), and the combined name and activity similarity. (c) ICL significantly increases hypothesis alignment of Llama2 embeddings for activities and combined hypothesis matrices, $p< 0.05$.}
    \label{fig:nah-beh}
\end{figure*}

\textbf{Hypothesis alignment.} We formed three hypothesis matrices about what the similarity structure should look like between prompt embeddings, and measured how close to our hypotheses the actual similarity matrices are. 
\begin{itemize}
    \item \textbf{Activity-based hypothesis}: We hypothesize if the correct answer to two prompts involves the same activity, then those prompts should have more similar embeddings compared to pairs that do not satisfy this condition. We construct the hypothesis matrix $H_{act}$ so that $H_{act}[i, j] = 1$ if prompts $i$ and $j$ have the same activity in their correct answer, and $0$ otherwise.\vspace{-2mm}
    \item \textbf{Name-based hypothesis}: We hypothesize if two prompts inquire about the same person, then those prompts should have more similar embeddings compared to pairs that do not satisfy this condition. We construct the hypothesis matrix $H_{name}$ so that $H_{name}[i, j] = 1$ if prompts $i$ and $j$ inquire about the same person, and $0$ otherwise.\vspace{-2mm}
    \item \textbf{Combined hypothesis}: We combine the above two hypotheses and expect prompt pairs that satisfy both conditions to be most similar ($H_{comb}[i, j] = 1$), those that satisfy only one condition to be less similar ($H_{comb}[i, j] = 0.5$), and those that satisfy none to be the least similar pairs ($H_{comb}[i, j] = 0$). 
\end{itemize}
\vspace{-3mm}
We computed the correlation between each hypothesis matrix and the prompt-to-prompt similarity matrix, and observed that when we add an ICL example to the prompt, the hypothesis alignment increases significantly for all hypotheses. This effect is statistically significant for all hypotheses in Vicuna, and for $H_{act}$ and $H_{comb}$ hypotheses in Llama-2 (Figures \ref{sub_fig:nah-rsa-vicuna-all-hyp} and \ref{sub_fig:nah-rsa-llama-all-hyp}).

We measured alignment with $H_{comb}$ for five layers of various depths in both models, and observed that the intermediate layers in Vicuna show consistent significant increase in hypothesis alignment after ICL (Figure \ref{sub_fig:nah-rsa-vicuna-comb-layers}), while in Llama-2, only the $60th$ layer shows statistically significant increase. We used the Fisher z-transformation to compare correlations before and after ICL and used $p < 0.01$ as the significance threshold.

\subsection{Attention Ratio Analysis of Composite Prompts.} 
\label{subsection:na_ara}
\begin{figure*}[!t]
    \centering
    \subfigure[Llama-2 ARA]{
        \includegraphics[width=.22\textwidth]{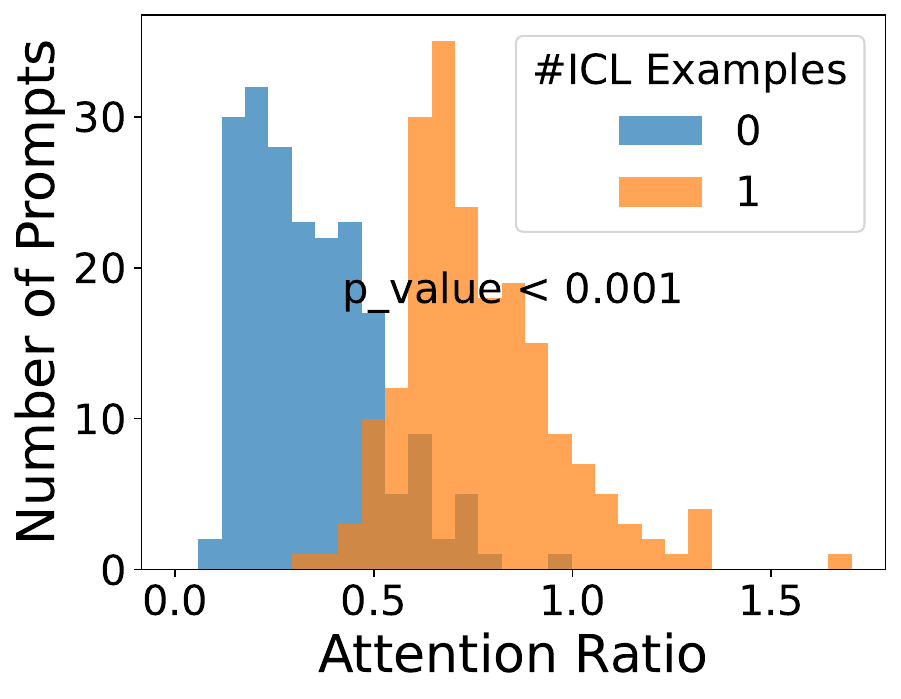}
        \label{sub_fig:attn-ratio-nah-llama}}
    \subfigure[Vicuna ARA]{
        \includegraphics[width=.22\textwidth]{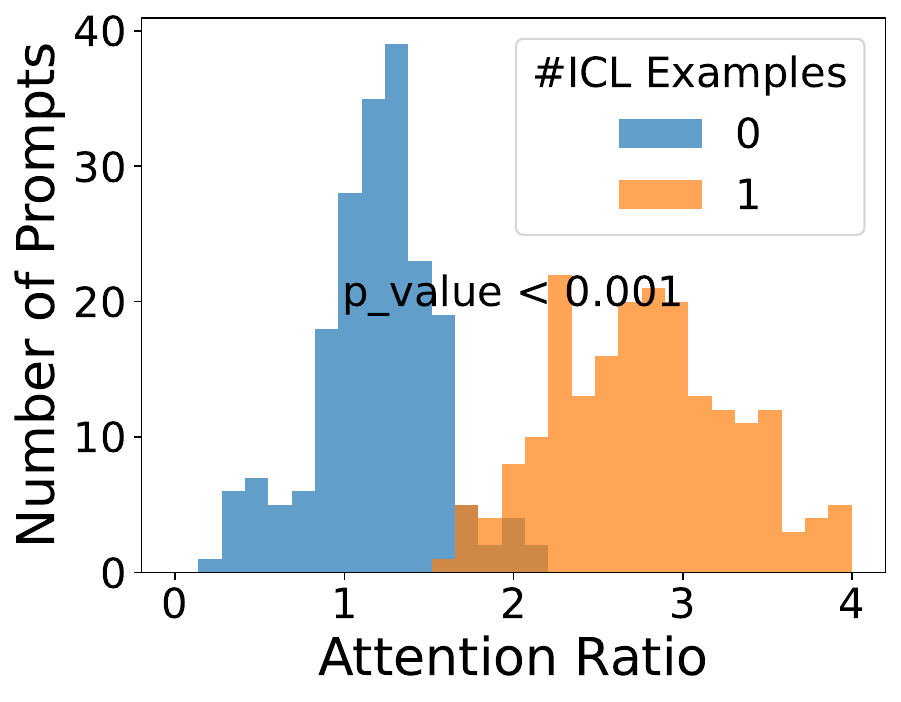}
        \label{sub_fig:attn-ratio-nah-vicuna}}
    \subfigure[Vicuna ARA vs behavior]{
        \includegraphics[width=.22\textwidth]{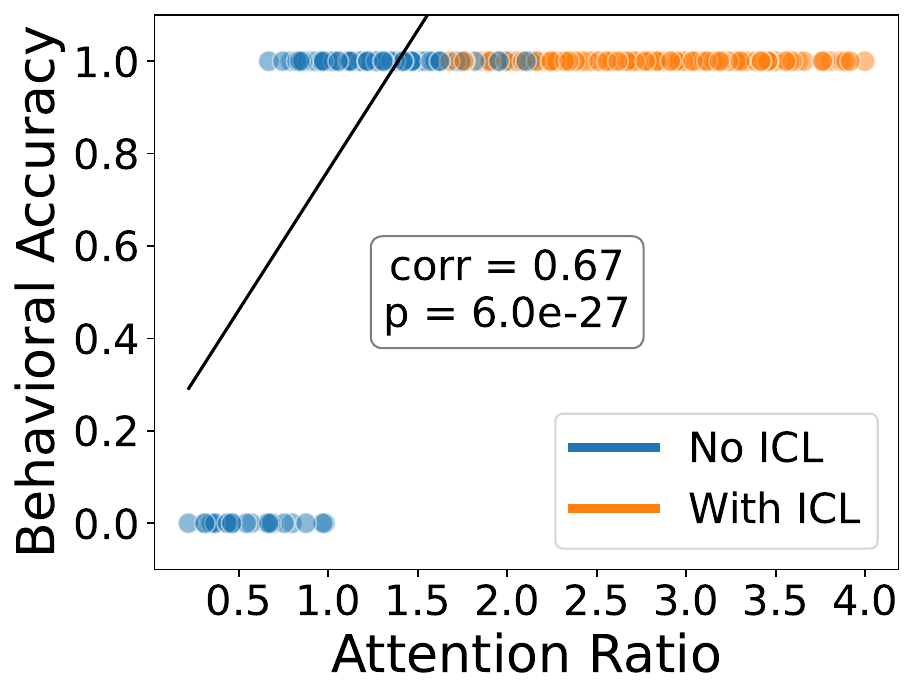}
        \label{sub_fig:vic_attn_vs_beh_middle}}
    \subfigure[Llama-2 
    ARA vs behavior]{
        \includegraphics[width=.22\textwidth]{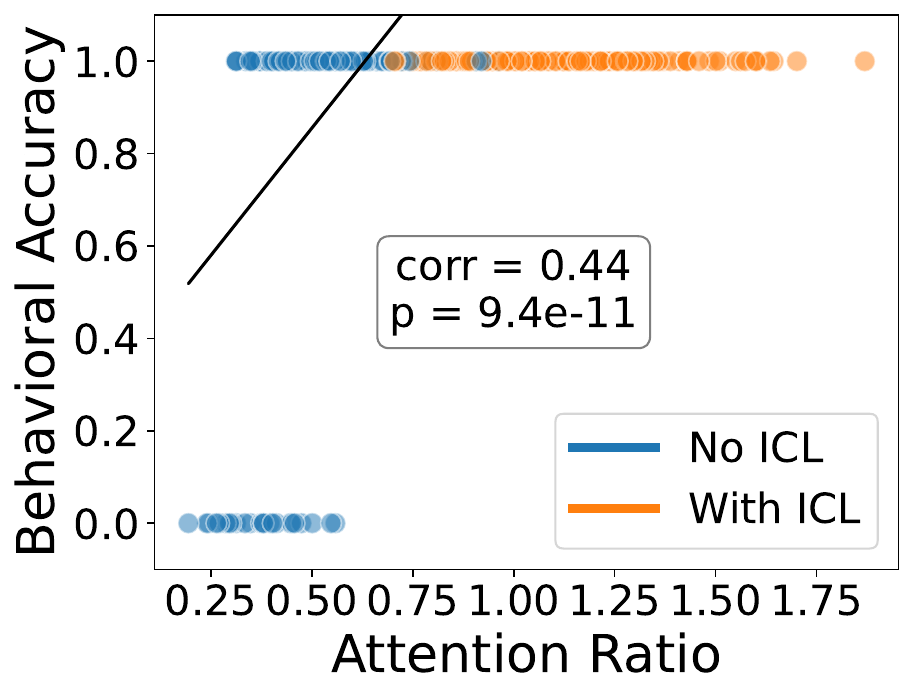}
        \label{sub_fig:llama_attn_vs_beh_middle}}        
    \caption{\textbf{Attention Ratio Analysis (ARA)  before and after ICL for the reading comprehension task.}. (a) and (b) The ratio of attention to informative and uninformative information were measured for the reading comprehension task before and after ICL (blue and orange respectively) for Llama-2 70B and Vicuna-1.3 13B. Attention ratio distributions concentrated toward larger numbers indicate more attention to the informative part of the prompt. Attention ratios corresponding to the middle layer of both models significantly shift to larger values with the introduction of ICL, indicating more attention to informative information after ICL. (c) and (d) Attention ratios (x axis) in both models are significant indicators of each model's behavioral accuracy (y axis).}
    \label{fig:attn-ratios-nah}
\end{figure*}

\begin{figure*}[!t] 
\centering
    \subfigure[Llama-2 ARA]{
        \includegraphics[width=.22\textwidth]{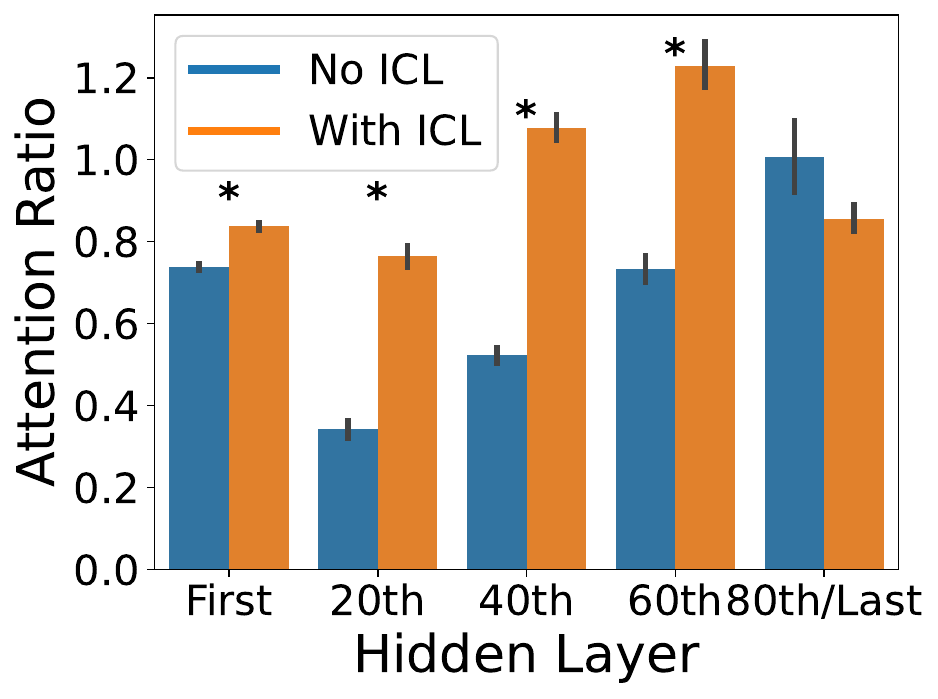}
        \label{sub_fig:attn-ratio-nah-llama-layers}}
    \subfigure[Vicuna ARA]{
        \includegraphics[width=.22\textwidth]{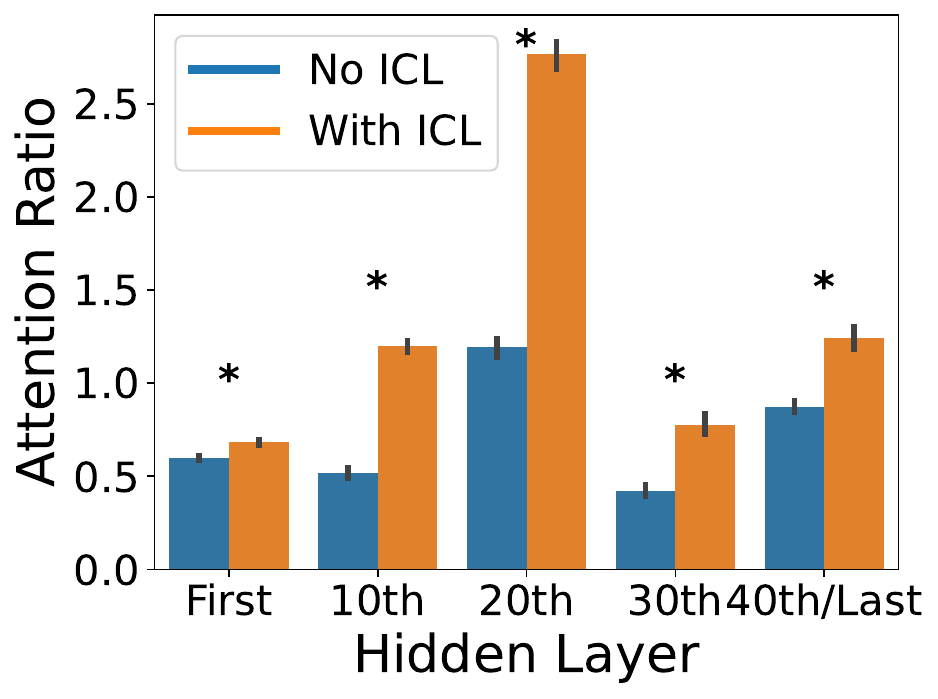}
        \label{sub_fig:attn-ratio-nah-vicuna-layers}}
   \subfigure[Vicuna alignment]{
        \includegraphics[width=.22\textwidth]{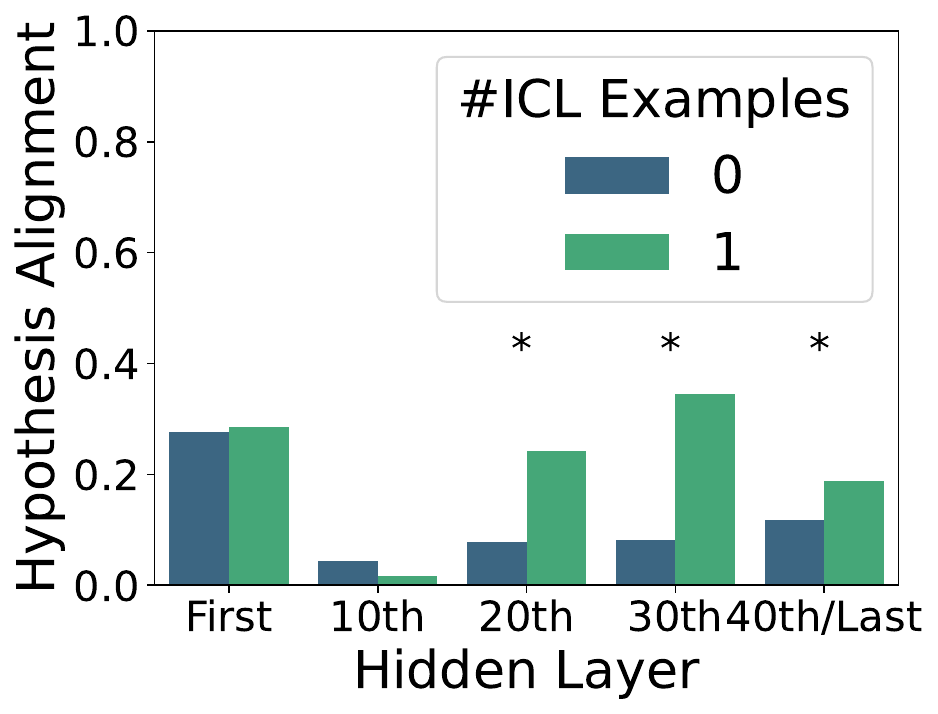}
        \label{sub_fig:nah-rsa-vicuna-comb-layers}}
    \subfigure[Llama-2 alignment]{
        \includegraphics[width=.22\textwidth]{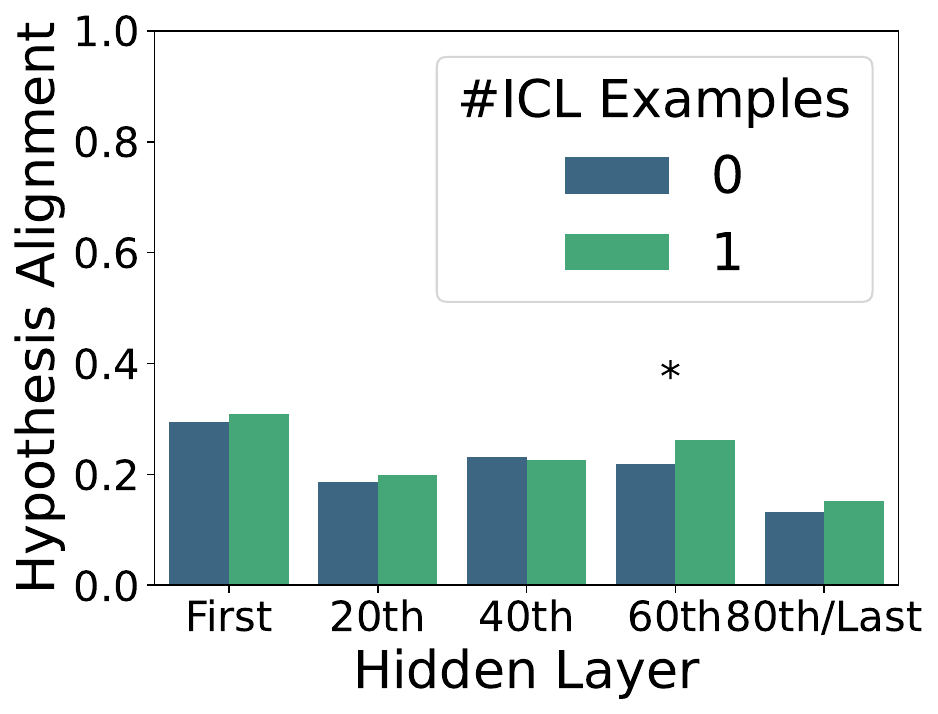}
        \label{sub_fig:nah-rsa-llama-comb-layers}}
    \vspace{-2mm}
    \caption{\textbf{Hypothesis alignment across LLM layers before and after ICL.} We measured the ratio of attention to informative vs. uninformative information, and hypothesis alignment of embeddings, across five layers (first, middle, last and quartiles) for each model.\ref{sub_fig:attn-ratio-nah-llama} All layers of Llama2, with the exception of the last layer, demonstrate a statistically significant increase in ratio of attention to informative parts of the prompt after ICL. \ref{sub_fig:attn-ratio-nah-vicuna} All five Vicuna layers demonstrate the same pattern significantly. * indicates statistically significant increase ($p < 0.001$). \ref{sub_fig:nah-rsa-vicuna-all-hyp}  Vicuna's embeddings demonstrate a significantly higher alignment with the combined hypothesis $H_{comb}$ in the middle and deeper layers. \ref{sub_fig:nah-rsa-llama-all-hyp} Llama's embeddings show this behavior across various layers although this effect is only statistically significant in layer $60$. * indicates statistical significance with $p < 0.01$.  }
    \label{fig:nah-rsa-layers}
\end{figure*}

Next we applied the attention ratio analysis described in Section \ref{subsection:attn} to composite prompts. Each composite prompt consists of well-defined informative ($s_{inf}$) and distracting simple prompts. For the example prompt above, ``Joseph is swimming." is a distracting simple prompt, while ``Patricia is reading a book" is an informative one from which the correct answer can be inferred. We analyzed the attention of LLMs' response $r$ on $s_{inf}$ to verify if an increased attention to the informative part of the prompt is underlying the behavior improvement after ICL. The simple prompts in composite prompts were shuffled before presenting the prompt to the models, so that the informative simple prompt does not always come first. For each composite prompt, we calculated the ratio of the response $r$ attention to $s_{inf}$ over response attention to the the whole prompt $p$ as $A(r, s_{inf}, p)$. In Figures \ref{sub_fig:attn-ratio-nah-llama} and \ref{sub_fig:attn-ratio-nah-vicuna}, we compare the distribution of this value over composite prompts before and after introduction of ICL in both models. The addition of one ICL example significantly shifts attention to the informative parts of the prompt (i.e., larger values of attention ratio). We consistently observe this ICL-induced improvement of the attention ratio distribution across both LLMs, in five different layers of various lengths (See figures \ref{sub_fig:attn-ratio-nah-vicuna-layers} and \ref{sub_fig:attn-ratio-nah-llama-layers}). Interestingly, the middle layers in both models show this effect most strongly. 

Importantly, the attention ratio is significantly correlated with the behavior of both LLMs (Figures \ref{sub_fig:vic_attn_vs_beh_middle} and \ref{sub_fig:llama_attn_vs_beh_middle}). Using mean-aggregation over attention heads instead of max-aggregation, and observed the same increase in attention ratios after ICL (See Appendix Figure \ref{fig:app_na_ara_mean}).

\section{Discussion and future directions}

We investigated how ICL improves LLM behavior, studying how it impacts embeddings and attention weights across layers. We designed a linear regression task and a reading comprehension task and tested two open source LLMs: Vicuna-1.3 (13B) and Llama-2 (70B). We analyzed changes in latent representations of these LLMs before and after ICL, measuring representational similarity among embeddings, hypothesis alignment, and attention ratios as well as their correlation with behavior. We found that ICL improves task-critical representations and attention allocation to relevant content in the prompt. These representational changes were correlated with behavioral gains.

\section{Related Work} 
Our use of RSA builds on prior work applying this technique to study neural representations in brains, neural networks, and NLP models. The latter includes examining relationships between sentence encoders and human processing \citep{abdouHigherorderComparisonsSentence2019}, correlating neural and symbolic linguistic structures \citep{chrupalaCorrelatingNeuralSymbolic2019}, analyzing biases in word embeddings \citep{leporiUnequalRepresentationsAnalyzing2020}, comparing word embedding and fMRI data \citep{fereidooniUnderstandingImprovingWord2020}, investigating encoding of linguistic dependencies \citep{leporiPickingBERTBrain2020}, and assessing semantic grounding in code models \citep{naikProbingSemanticGrounding2022}. 

Uniquely, we apply RSA at scale to study ICL in large pretrained models like Llama-70B. Our work shows RSA can provide insights into task-directed representation changes during ICL. Concurrent lines of work also aim to elucidate ICL mechanisms. One hypothesis is that Transformers implement optimization implicitly through self-attention. For example, research shows linear self-attention can mimic gradient descent for regression tasks \citep{vonoswaldTransformersLearnIncontext2022}. Other studies also suggest Transformers can implement gradient descent and closed-form solutions given constraints \citep{akyurekWhatLearningAlgorithm2022}, and specifically mimic preconditioned gradient descent and advanced techniques like Newton's method \citep{ahnTransformersLearnImplement2023}. However, to our knowledge ours is the first study to use RSA at scale, studying ICL in large language models trained on naturalistic data rather than toy models. 

This is important since insights from formal studies analyzing small toy models may not transfer to large pretrained models. A key advantage of our approach is the focus on larger LLMs like Llama-2: by scaling RSA and attention analysis our approach offers explanatory insights into real-world ICL capabilities. Our results show ICL improves embedding similarity according to experimental design (i.e., embeddings for tasks with cognitive similarity become more similar to each other), and shifts attention to relevant information, which also increases robustness to distractors. This aligns with the view that ICL relies on implicit optimization within the forward pass \citep{akyurekWhatLearningAlgorithm2022,ahnTransformersLearnImplement2023}. Moreover, the changes we observe in representations and attention after more ICL examples imply the model optimizes its processing of prompts in context.

% Some studies also model ICL through a Bayesian lens, viewing pretraining as learning a latent variable model for conditioning on prompts. For example, research posits LLMs marginalize out latent factors in prompts for conditioned predictions \citep{xieExplanationIncontextLearning2022}. Other work characterizes LLMs as implicit topic models inferring latent concepts from prompts \citep{wangLargeLanguageModels2023}, or as Bayesian predictors adapting via data distributions \citep{ahujaInContextLearningBayesian2023}. From this perspective, pretraining can be viewed as learning distributions over downstream tasks \citep{wiesLearnabilityInContextLearning2023}.

Relatedly, some studies model ICL through a Bayesian lens, viewing pretraining as learning a latent variable model for conditioning \citep{xieExplanationIncontextLearning2022, wangLargeLanguageModels2023, ahujaInContextLearningBayesian2023, wiesLearnabilityInContextLearning2023}. We empirically demonstrate that prompt embeddings become more task-aligned and attention more focused on critical task information. These observable changes could provide some additional support for the view that LLMs are effectively conditioning on salient factors implicit in prompts. In this sense, our results provide complementary real-world empirical evidence at the level of representations to supplement the theoretical insights from probabilistic perspectives.

The emerging field of mechanistic interpretability aims to reverse engineer model computations, drawing analogies to software decompiling. The goal is to recover human-interpretable model descriptions in terms of learned ``circuits'' implementing meaningful computations. For instance, recent work presents evidence that ``induction heads'' are a key mechanism enabling ICL in Transformers, especially in small models \citep{olsson2022context}. While mechanistic interpretability is promising for validating causal claims, it remains challenging to scale up. Automating circuit discovery is an active area \citep{conmy2023towards}, but not yet viable for models with tens of billions of parameters. Our approach provides complementary evidence by showing how relevant information becomes encoded in embeddings and attention. While we do not isolate causal circuits, we demonstrate the behavioral effect of improved task representations. Thus, we believe our proposed application of RSA and attention ratios could help evaluate proposals from mechanistic research in the future.

LLMs have been shown to fail at multi-step planning and reasoning  \citep{cogeval,allure}. A future direction is to study the effects of ICL on improving planning behavior in LLMs. Specifically, analyzing the latent representations of the different layers before and after ICL, and measuring the correlation between changes in representations and improvements in planning behavior on Markov decision processes.

In sum, we show that ICL improves LLM behavior by better aligning its embedding representations and attention weights with task-relevant information. In future work, we intend to apply the method to better understand how LLMs work, and implement the methods offered here as a white-box augmentation of LLMs.

\section{Impact Statement}
% Given we have access to embeddings and attention weights of open-source LLMs, neuroscience-inspired methods can offer insights into the inner-workings of LLMs in relation to expected functions, with or without ICL or chain of thought commands, etc. This paper presents work inspired by neuroscience methods with the goal of advancing the field of Machine Learning. We foresee positive potential societal consequences. Offering a better understanding of the internal representations of LLMs and quantifying their alignment with what they are expected to do, our approach evaluates the interpretability, safety, and improvability of LLMs and alignment with  intended functions.

Given that we have access to embeddings and attention weights of open-source LLMs, neuroscience-inspired methods can offer insights into the inner workings of LLMs in relation to expected functions, with or without ICL or chain of thought methods. This paper presents work inspired by neuroscience methods with the goal of advancing the field of machine learning. 

A cornerstone of our approach is the demonstration that interpretability methods inspired by neuroscience, such as RSA, can be effectively scaled to LLMs with tens of billions of parameters. As a result, such methods can be deployed on state-of-the-art LLMs, moving beyond the constraints of smaller toy models often used in interpretability research. Our work, therefore, bridges the gap between theoretical research and practical application, offering a path to scrutinize and understand the complex behaviors of production-level LLMs.

The capacity to analyze and predict LLM behavior also offers potential societal benefits, such as protecting against misuse. For example, gaining a better understanding of the representational shift underpinning ICL is a meaningful step towards detecting and preventing prompt injection attacks.

Finally, our approach demonstrates the value of interdisciplinary research, bridging insights from neuroscience and machine learning. This cross-pollination not only enriches our understanding of artificial systems but may also offer back novel insights for the study of neural representations in the brain. We encourage further exploration along these interdisciplinary lines to foster mutual advancements in AI and computational neuroscience.

\bibliography{iclr2024_conference}
\bibliographystyle{iclr2024_conference}
\clearpage

\appendix
% \section{Appendix}

% \section{Additional Experimental Details}

\section{Linear Regression}
\label{sec:appendix_regression}

\subsection{Decodability Results}
\label{app:reg_probe}
\begin{figure*}
    \centering
    \subfigure[Llama2 probe using max-aggregation]{
    \includegraphics[width=0.3\textwidth]{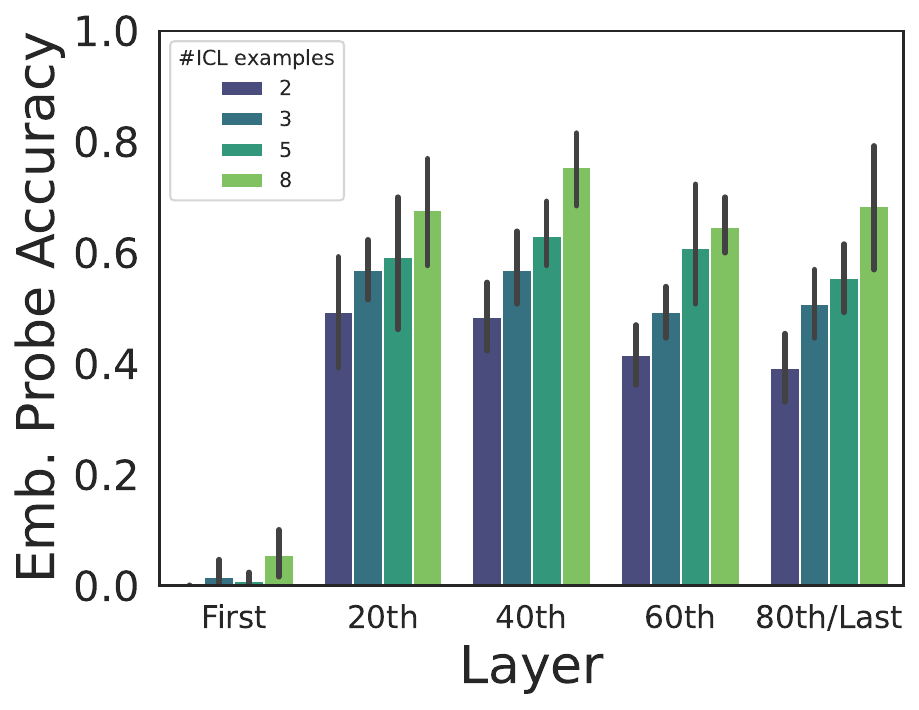}
    \label{subfig:reg_prob_a}
    }
    \subfigure[Vicuna probe using max-aggregation]{
    \includegraphics[width=0.3\textwidth]{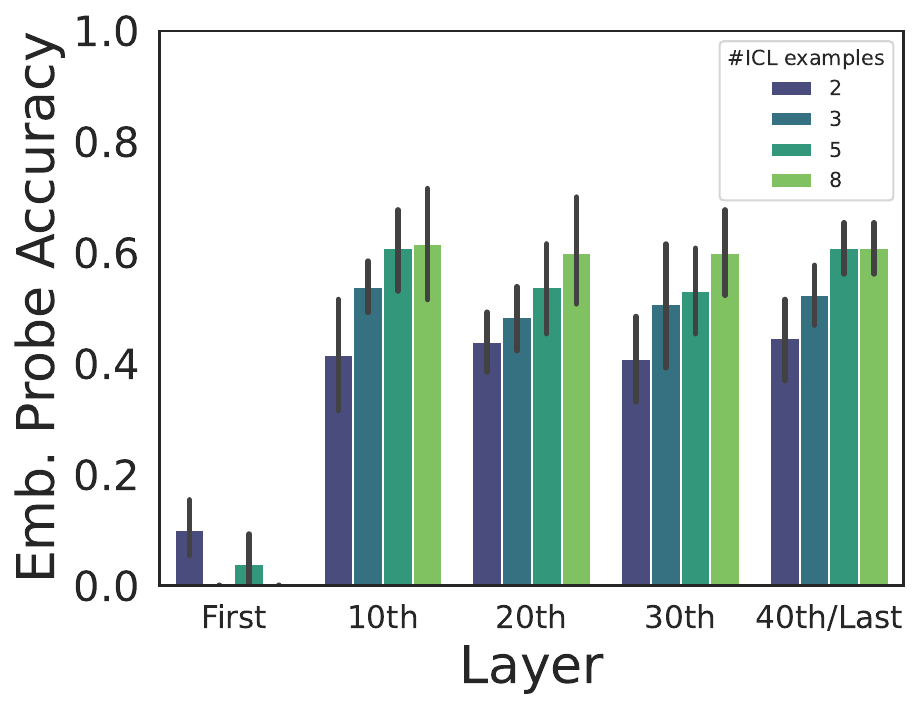}
    \label{subfig:reg_prob_b}
    }
    \subfigure[Vicuna probe using mean-aggregation]{
    \includegraphics[width=0.3\textwidth]{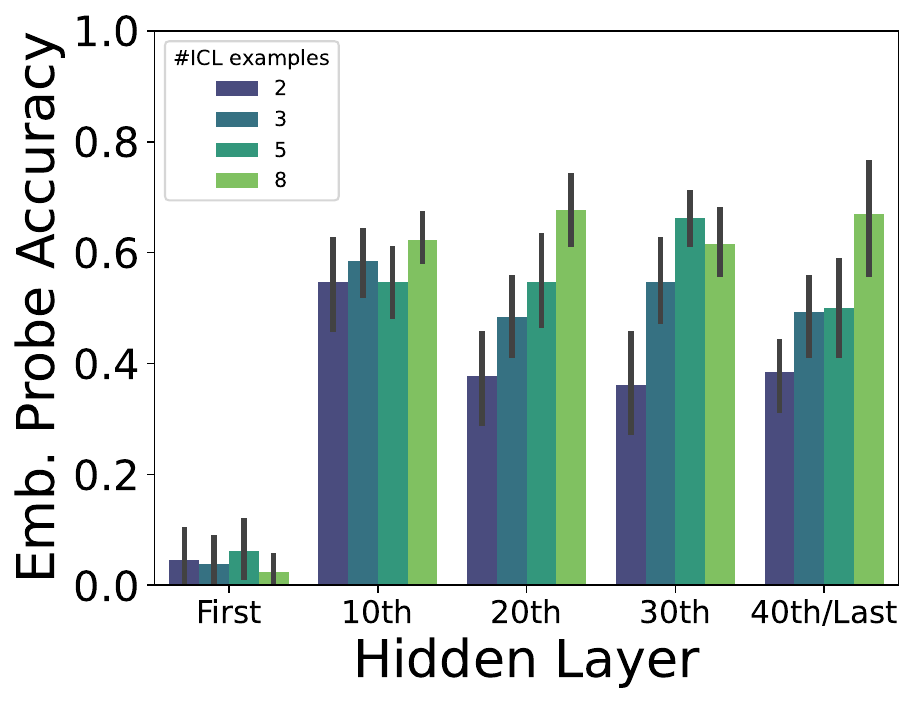}
    \label{subfig:reg_prob_c}
    }
    \caption{Accuracy of probing classifiers trained on (a) max-aggregated Llama2 embeddings, (b) max-aggregated Vicuna embeddings, (c) and mean-aggregated Vicuna embeddings to predict the line slope in the linear regression task, using prompt embeddings taken from various hidden layers. There is a consistent improvement to the probes' performance as we add more ICL example points to the prompts. This pattern is consistent across models, layers, and embedding aggregation methods.}
    \label{fig:app_reg_probe_slope_layers}
\end{figure*}

 In Section \ref{regression_exps}, we studied the ability of probing classifiers to decode the line slope from the prompt embeddings and showed that the accuracy of these classifiers increases as we increase the number of ICL examples (see Figure \ref{sub_fig:reg_emb_probe}). Here, we extend that experiment to several layers of various depths in both models. We observed a consistent pattern of improvement in probe accuracy in all layers except the first one. See Figure \ref{fig:app_reg_probe_slope_layers}. 

 Additionally, we repeated this experiment with prompt embeddings obtained by mean-pooling over token embeddings instead of max-pooling and observed similar results indicating more task-critical information (the line's slope) was encoded in the LLMs' latent representations after ICL examples were added to the prompts. See Figure \ref{subfig:reg_prob_c}.

 \subsection{RSA: Hypothesis Alignment}
\begin{figure*}
    \centering
    \subfigure[Vicuna RSA with mean-aggregation in Linear Regression]{
    \includegraphics[width=0.4\textwidth]{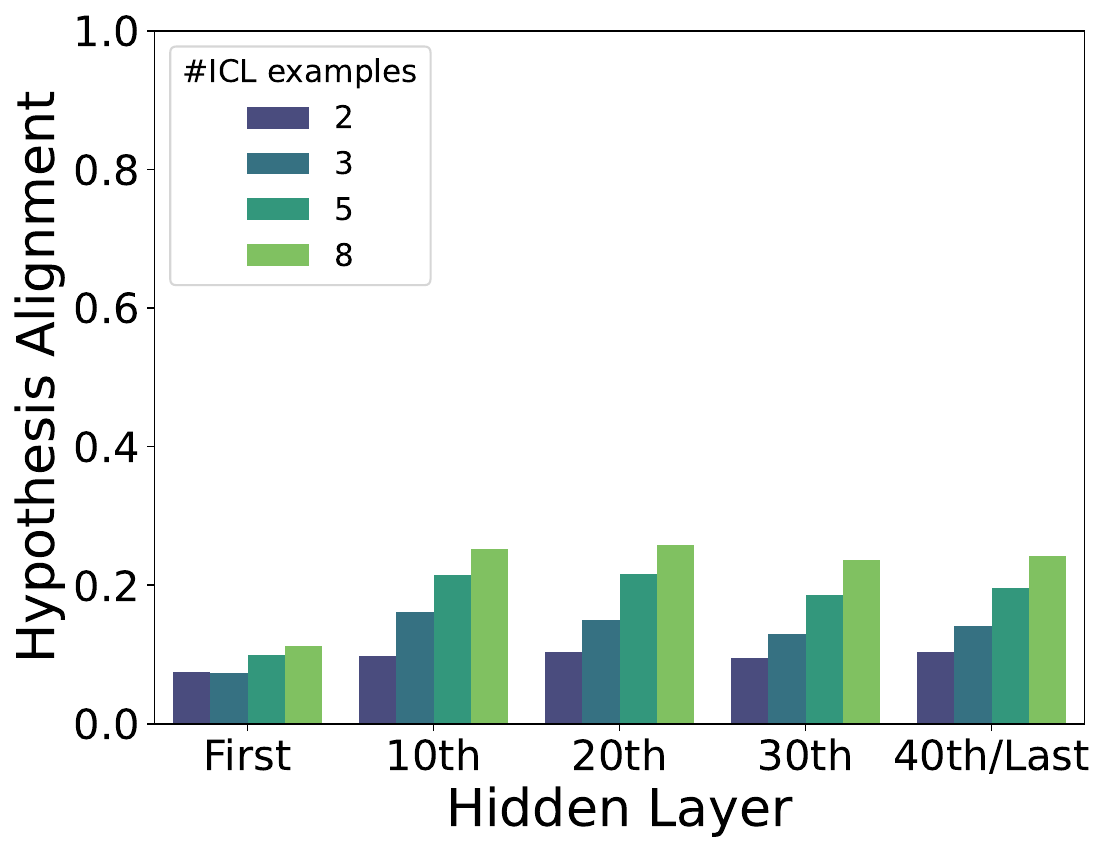}
    \label{subfig:reg_rsa_mean}
    }
    \subfigure[Llama-2 RSA with mean-aggregation in Linear Regression]{
    \includegraphics[width=0.4\textwidth]{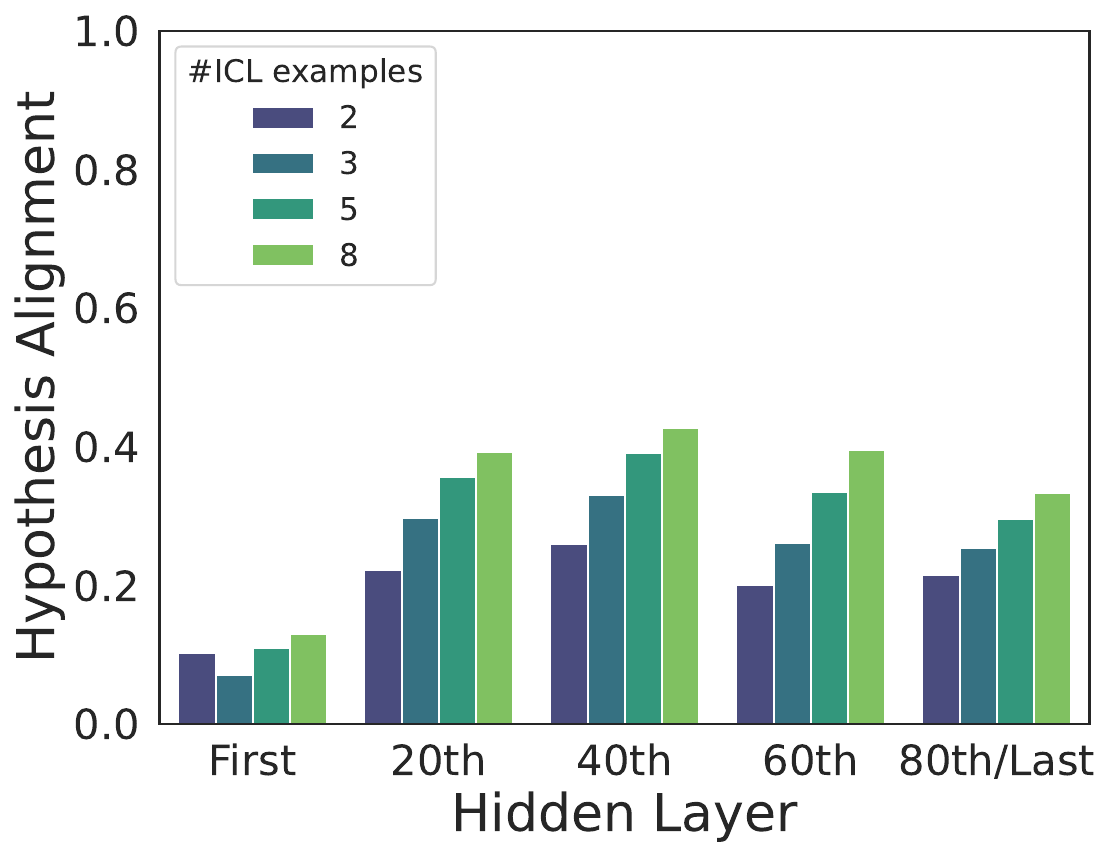}
    \label{subfig:reg_rsa_mean}
    }
    \caption{We measured alignment of the the linear regression prompts' embedding similarity matrix with the slope-based hypothesis matrix using mean-pooling over prompt token embeddings instead of max-pooling. Although with mean-pooling the hypothesis alignment was generally lower, we still observed increased alignment as we increased the number of ICL examples in the prompts in both Vicuna1.3 13B (a) and Llama-2 70B (b).}
    \label{fig:app_reg_rsa}
\end{figure*}

 We measured alignment of the linear regression prompts' embedding similarity matrix with the slope-based hypothesis matrix using mean-pooling over prompt token embeddings instead of max-pooling. We still observed increased alignment as we increased the number of ICL examples in the prompts in both Vicuna1.3 13B (a) and Llama-2 70B (b), although with mean-pooling the hypothesis alignment was generally lower for Vicuna. See Figure \ref{fig:app_reg_rsa}.

\section{Reading Comprehension: Names and Activities }
\label{sec:appendix_names_acts}

\begin{figure*}
    \centering
    \subfigure[Vicuna - ARA - mean aggregation]{
    \includegraphics[width=0.3\textwidth]{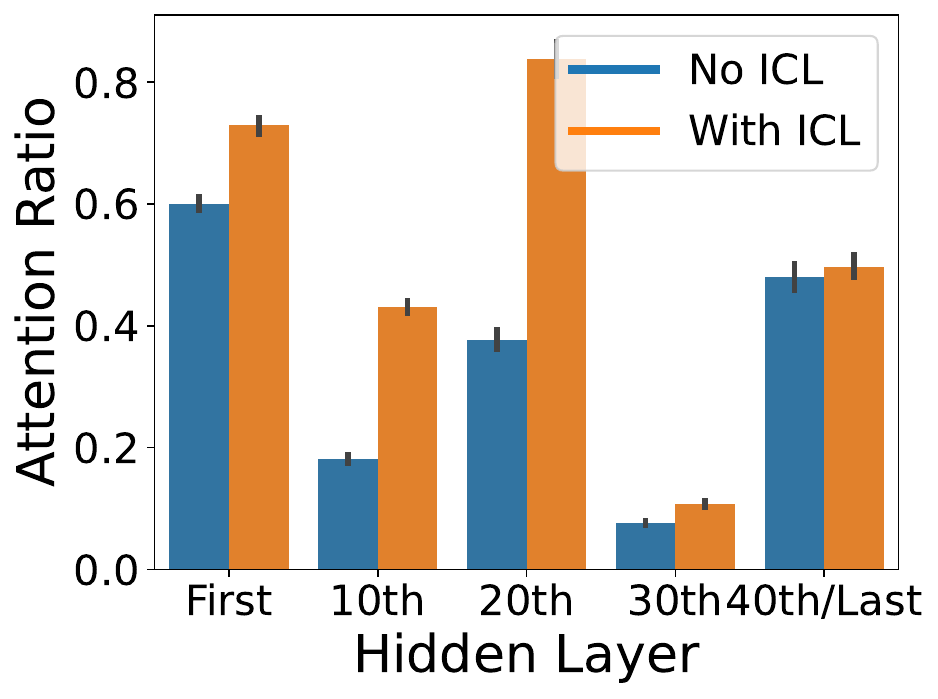}
    \label{subfig:app_ara_a}
    }
    \subfigure[Vicuna - ARA - mean aggregation]{
    \includegraphics[width=0.3\textwidth]{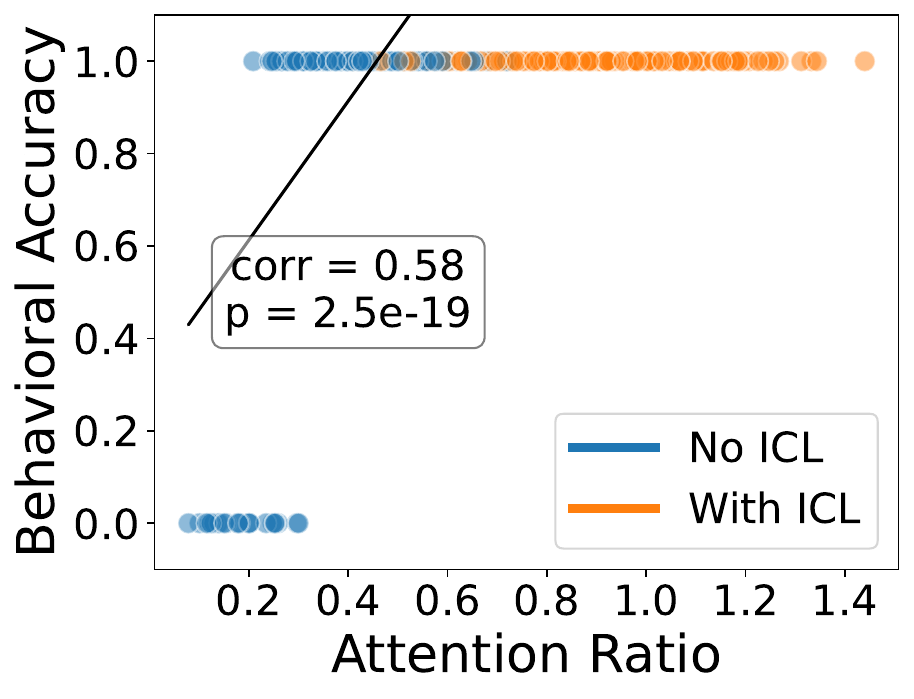}
    \label{subfig:app_ara_b}
    }
    \subfigure[Vicuna - ARA - mean aggregation]{
    \includegraphics[width=0.3\textwidth]{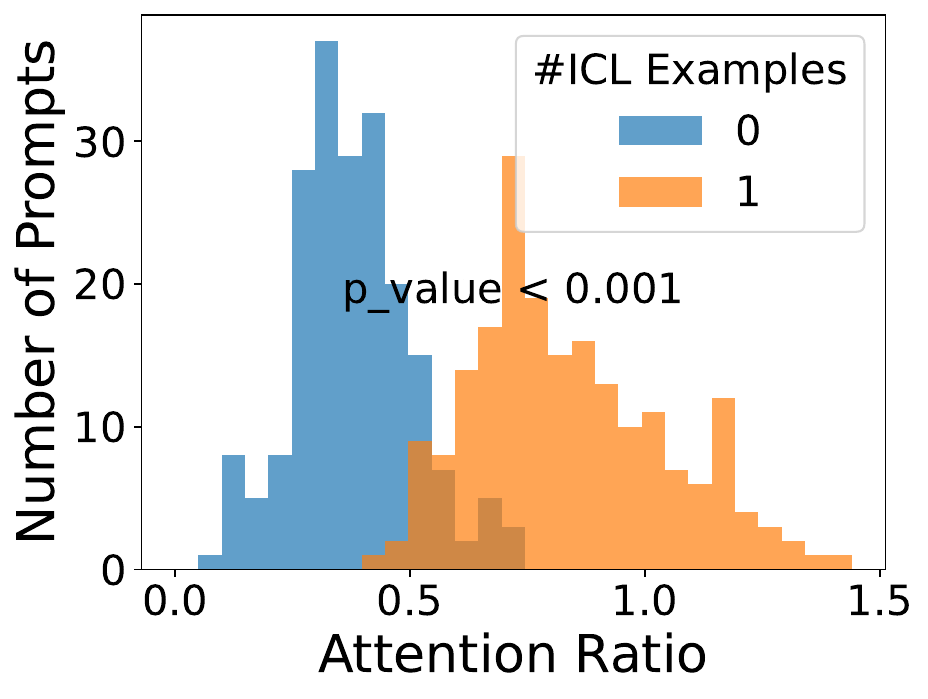}
    \label{subfig:app_ara_c}
    }
    \subfigure[Llama2 - ARA - mean aggregation]{
    \includegraphics[width=0.3\textwidth]{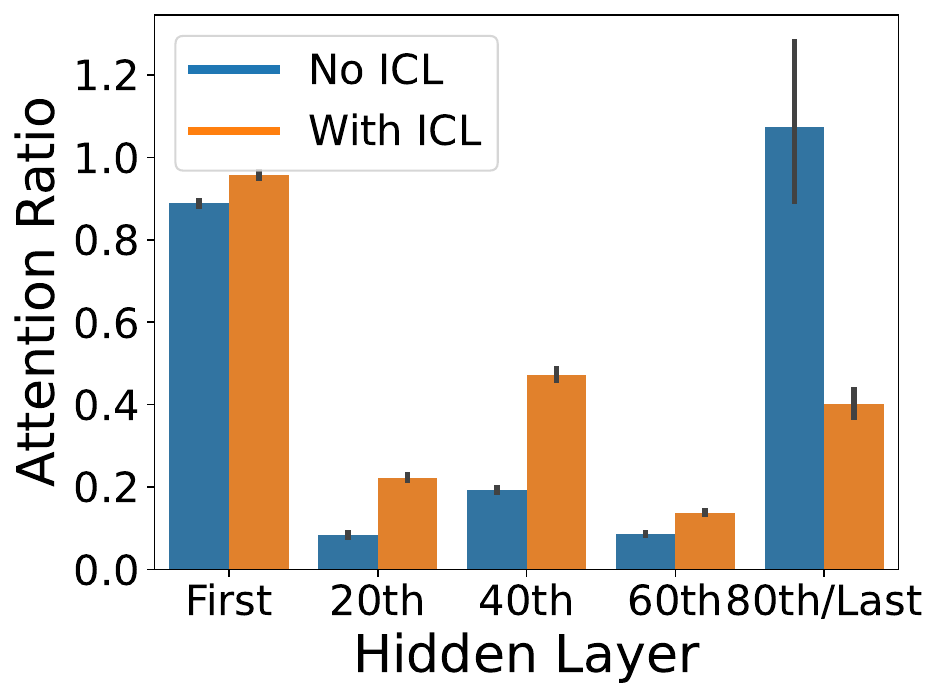}
    \label{subfig:app_ara_d}
    }
    \subfigure[Llama2 - ARA - mean aggregation]{
    \includegraphics[width=0.3\textwidth]{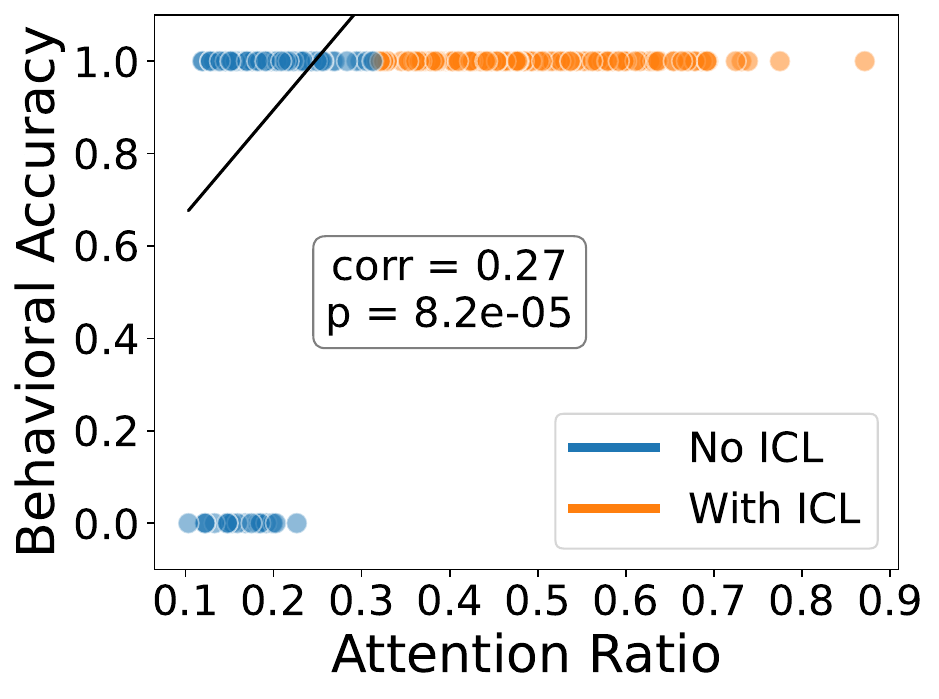}
    \label{subfig:app_ara_e}
    }
    \subfigure[Llama2 - ARA - mean aggregation]{
    \includegraphics[width=0.3\textwidth]{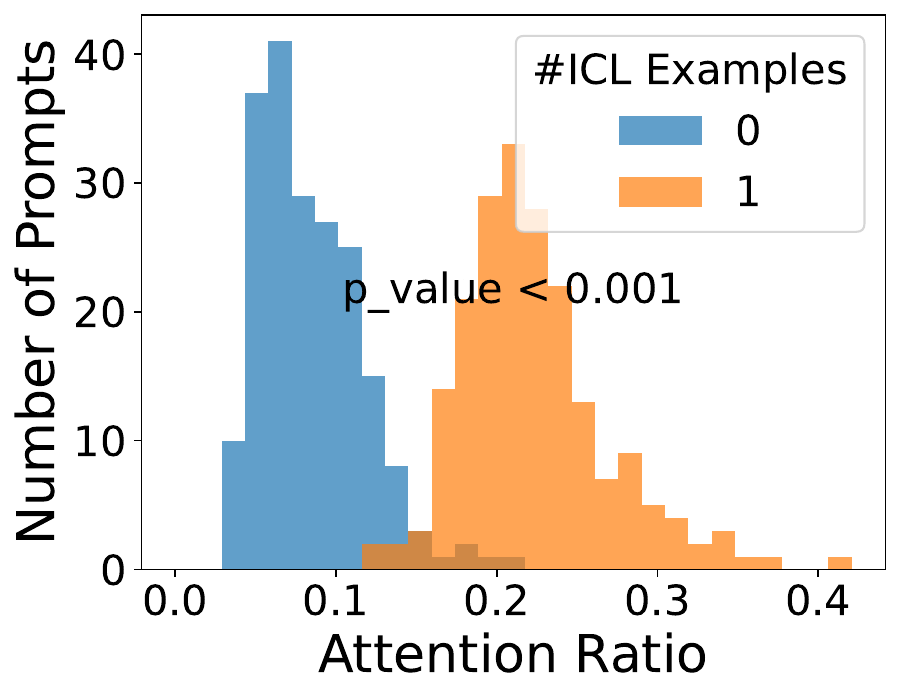}
    \label{subfig:app_ara_f}
    }
    \caption{\textbf{Attention Ratio Analysis (ARA) using mean aggregated attention weights over all attention heads.} In the paper, we report ARA results using attention weights that were max-aggregated over attention heads. Here we verify if the results still hold using a different aggregation method.  \textbf{(a)and(d)} The attention ratios significantly ($p < 0.01$) increase in all layers of both models except the last layer. \textbf{(b)and(e)} Attention ratios calculated from mean-aggregated attention weights are significantly correlated with the correctness of model response. \textbf{(c)and(f)} We show the distribution of attention ratios calculated from mean-aggregated attention weights of the middle layer of both models. A statistically significant to larger rations indicates more attention paid to the relevant content of the prompt. (See Section \ref{subsection:attn})}
    \label{fig:app_na_ara_mean}
\end{figure*}

\subsection{Prompt Samples}
\label{sec:appendix_names_acts_sample_prompts}
Here we show examples of composite prompts used in Sec~\ref{Exp1} for the reading comprehension task. Our goal was to study
the underlying changes in latent states and attention patterns that lead to performance improvement in face of distractors, so it was important to find tasks that were challenging for the models without ICL, and that could be made easier with ICL. We found that different distractors pose a challenge to Llama-2 and Vicuna-1.3 where ICL can help:

\subsubsection{Llama-2}
Sample prompts for \textbf{Llama-2} with and without in-context examples:

\textbf{Without ICL}
\begin{framed}
Oliver is baking a cake. Joseph is playing basketball. Ileana is sleeping. Natalie is doing the same thing as Ileana. Natalie carpools to work with Joseph. Natalie enjoys playing basketball with Joseph. What is Natalie doing?
\end{framed}

\textbf{With ICL}
\begin{framed}
Question: Bob is playing the piano.  Xavier is playing basketball. Tina is doing the same thing as Bob. Tina is a neighbor of Xavier. Tina enjoys rock climbing with Xavier. What is Tina doing? Answer: Tina is playing the piano. Question: Ileana is riding a bike. Harriet is watching TV. Joseph is doing the same thing as Harriet. Joseph is best friends with Ileana. Joseph enjoys riding a bike with Ileana. What is Joseph doing? Answer:	
\end{framed}

\subsubsection{Vicuna-1.3}
Sample prompts we used for \textbf{Vicuna-1.3} with and without in-context examples:

\textbf{Without ICL}
\begin{framed}
Patricia is baking a cake. Harriet is singing a song. Natalie is doing the same thing as Harriet. Natalie is a coworker of Patricia. What is Natalie doing?
\end{framed}

\textbf{With ICL}
\begin{framed}
Question: Alice is cooking a meal. Ursula is singing a song. Wendy is doing the same thing as Alice. Wendy carpools to work with Ursula. What is Wendy doing? Answer: Wendy is cooking a meal. Question: Larry is swimming. Mark is riding a bike. Natalie is doing the same thing as Larry. Natalie is a neighbor of Mark. What is Natalie doing? Answer:	
\end{framed}

\subsection{Pairwise similarity matrix}
\label{sec:appendix_names_acts_sim}

\begin{figure*}
    \centering
    \subfigure[Simple prompt embeddings pairwise similarity matrix]{
    \includegraphics[width=\textwidth]{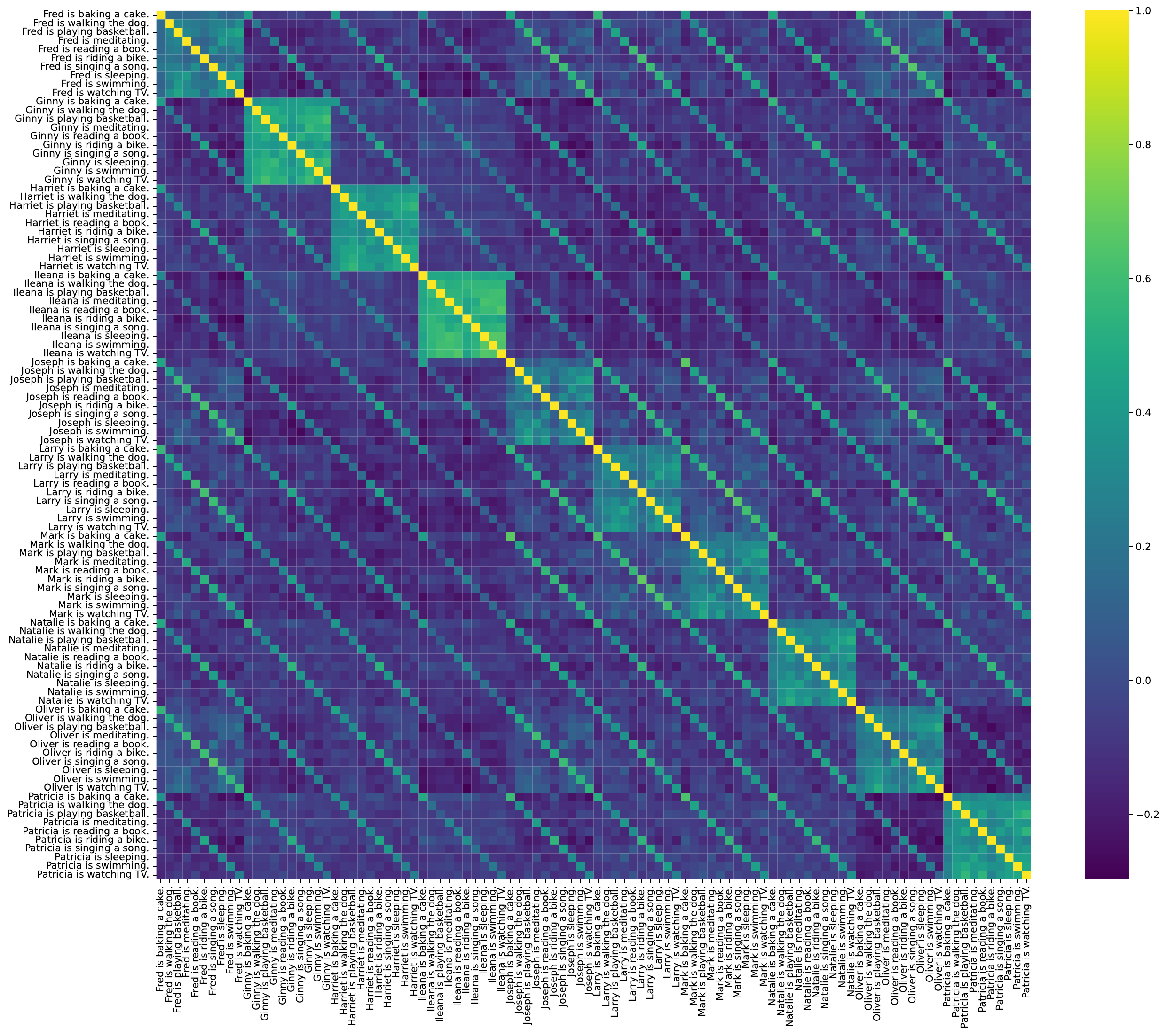}
    \label{fig:names_act_heatmap}}
    \subfigure[t-SNE: Llama-2 Embs]{
    \includegraphics[width=.33\textwidth]{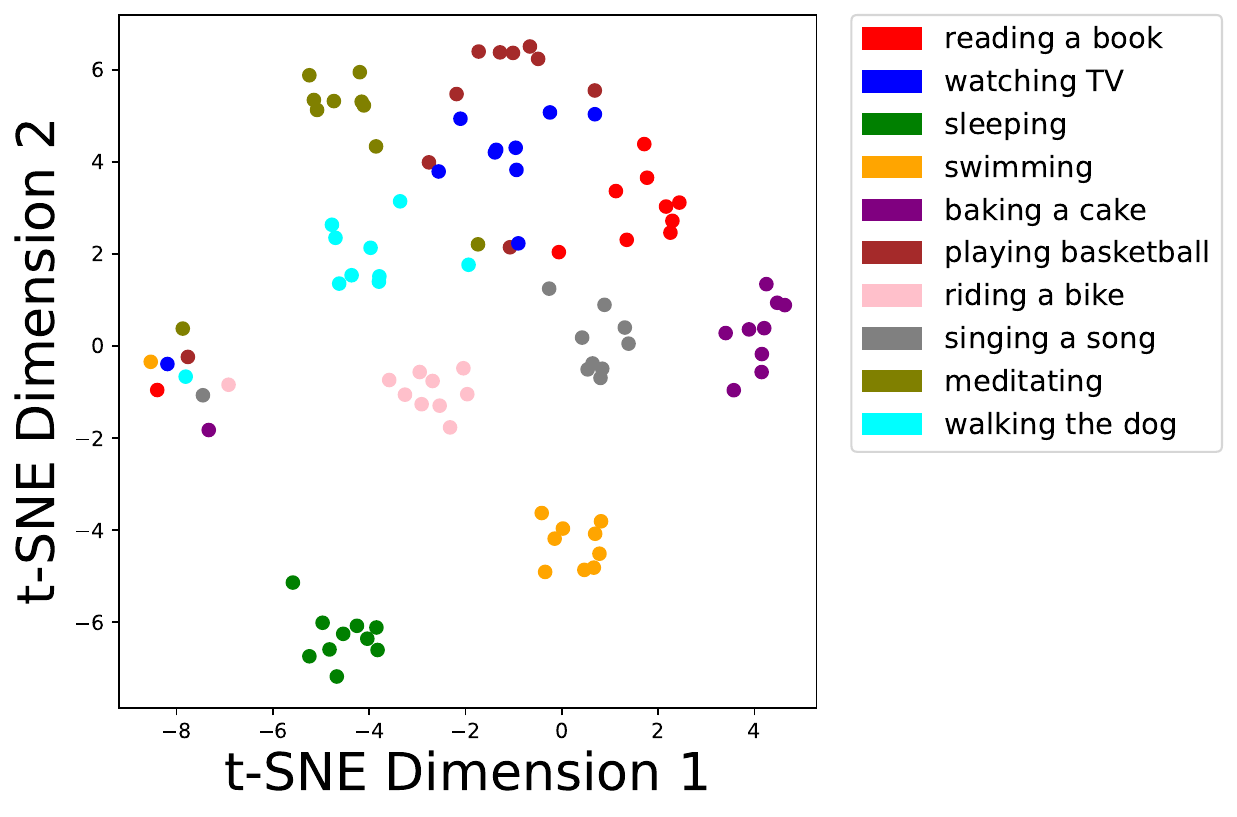}
    \label{fig:atomic-tsne}
    }
    \subfigure[Prompt Emb. Classifier]{
    \includegraphics[width=.28\textwidth]{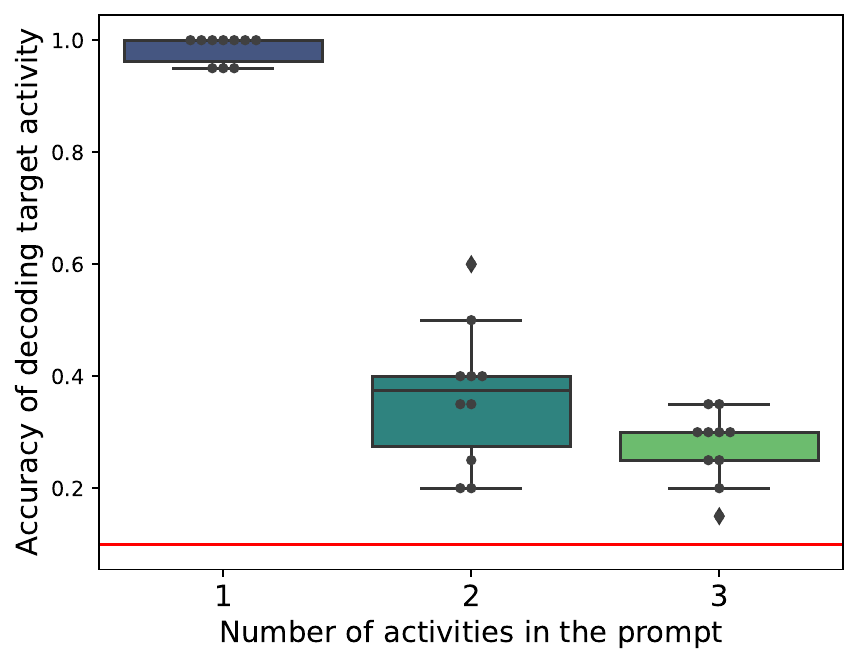}
    \label{fig:composite-decode}
    }
\caption{\textbf{Demonstration of Llama2 embedding analysis for the reading comprehension prompts (names and activities)}. (a) Pairwise embedding similarity matrix reveals a structure where simple prompts with the same name (bright squares), and prompts with the same activity (diagonal lines) are highly similar. For readability, only a subset of the prompts is included in this plot. See supplementary material for full set. (b) t-SNE visualization of the simple prompt embeddings reveals that embeddings of prompts describing the same activity are clustered together. Each point corresponds to a simple prompt and is colored by the activity mentioned in the prompt. (c) Accuracy of a logistic regression classifier trained to decode target activity from the embeddings of composite prompts that include one, two, or three simple prompts/activities and indirectly ask about one activity (target activity). The red line indicates random guessing accuracy. The task of predicting the target activity from prompt embeddings becomes harder as we add more simple prompts to the composite prompt.}
\end{figure*}

In this section we provide embedding visualization and classification for the prompts introduced in Section \ref{Exp1}. The pairwise similarity matrix of all simple prompt embeddings extracted from Llama-2 reveals high similarity between simple prompts involving the same name (bright blocks) and the same activity (bright diagonal lines). See Figure \ref{fig:names_act_heatmap}. Furthermore, t-SNE visualization of the same embeddings reveals that embeddings of prompts describing the same activity are clustered together. Each point corresponds to a simple prompt and is colored by the activity mentioned in the prompt. Most clusters in the t-SNE visualization homogeneously represent an activity (same color), one exception to this is the cluster on the left that corresponds to a specific name (``Ileana").

In order to validate the probing classifier approach, we trained a logistic regression classifier to classify the ground truth activity from the embeddings of composite prompts that include one, two, or three simple prompts. See Figure \ref{fig:composite-decode}. The task of predicting the ground truth activity from prompt embeddings becomes harder as we add more simple prompts to the composite prompt, which aligns with deteriorating behavioral accuracy (See Figure \ref{fig:composite-decode}).

\end{document}